\documentclass[conference]{IEEEtran}

\IEEEoverridecommandlockouts
\usepackage{times}
\usepackage{cite}
\usepackage[numbers]{natbib}
\usepackage{multicol}
\usepackage{hyperref}
\usepackage{amsmath,amssymb,amsfonts,amsthm}
\usepackage{algorithm}
\usepackage{algpseudocode}
\usepackage{graphicx}
\usepackage{textcomp}
\usepackage{xcolor}
\usepackage{subcaption}
\usepackage{dblfloatfix}
\usepackage{bbm}

\usepackage{mymacros}

\newtheorem{theorem}{Theorem}[section]

\begin{document}

\title{Certifiable Robot Design Optimization using Differentiable Programming
\thanks{C. Dawson is supported by the NSF GRFP under Grant No. 1745302. The Defense Science and Technology Agency in Singapore and IBM provided funds to assist the authors with their research, but this article solely reflects the opinions and conclusions of its authors and not DSTA Singapore, the Singapore Government, or IBM.}
}
\pdfinfo{
   /Author (Charles Dawson, Chuchu Fan)
   /Title  (Certifiable Robot Design Optimization using Differentiable Programming)
   /CreationDate (D:20220101120000)
   /Subject (Robot Design)
   /Keywords (robustness analysis, design optimization, automatic design tools)
}

\author{\IEEEauthorblockN{Charles Dawson}
\IEEEauthorblockA{\textit{Dept. of Aeronautics and Astronautics} \\
\textit{Massachusetts Institute of Technology}\\
Cambridge, USA \\
\texttt{cbd@mit.edu}}
\and
\IEEEauthorblockN{Chuchu Fan}
\IEEEauthorblockA{\textit{Dept. of Aeronautics and Astronautics} \\
\textit{Massachusetts Institute of Technology}\\
Cambridge, USA \\
\texttt{chuchu@mit.edu}}
}

\maketitle

\begin{abstract}
There is a growing need for computational tools to automatically design and verify autonomous systems, especially complex robotic systems involving perception, planning, control, and hardware in the autonomy stack. Differentiable programming has recently emerged as powerful tool for modeling and optimization. However, very few studies have been done to understand how differentiable programming can be used for robust, certifiable end-to-end design optimization. In this paper, we fill this gap by combining differentiable programming for robot design optimization with a novel statistical framework for certifying the robustness of optimized designs.
Our framework can conduct end-to-end optimization and robustness certification for robotics systems, enabling simultaneous optimization of navigation, perception, planning, control, and hardware subsystems.

Using simulation and hardware experiments, we show how our tool can be used to solve practical problems in robotics. First, we optimize sensor placements for robot navigation (a design with 5 subsystems and 6 tunable parameters) in under 5 minutes to achieve an 8.4x performance improvement compared to the initial design. Second, we solve a multi-agent collaborative manipulation task (3 subsystems and 454 parameters) in under an hour to achieve a 44\% performance improvement over the initial design. We find that differentiable programming enables much faster (32\% and 20x, respectively for each example) optimization than approximate gradient methods. We certify the robustness of each design and successfully deploy the optimized designs in hardware. An open-source implementation is available at \url{https://github.com/MIT-REALM/architect}.
\end{abstract}

\IEEEpeerreviewmaketitle

\section{Introduction}

To design complex systems, engineers in many fields use computer-aided tools to boost their productivity. Mechanical engineers can use a suite of 3D CAD (computer-aided design) and FEA (finite-element analysis) tools to design structures and understand their performance. Likewise, electrical engineers use electronic design automation tools, including hardware description languages like Verilog, to design and analyze large-scale, reliable, and yet highly complex integrated circuits. Sadly, when it comes to designing autonomous systems and robots, engineers often take an ad-hoc approach, relying heavily on experience and tedious parameter tuning.

\begin{figure}[tb]
    \centering
    \includegraphics[width=\linewidth]{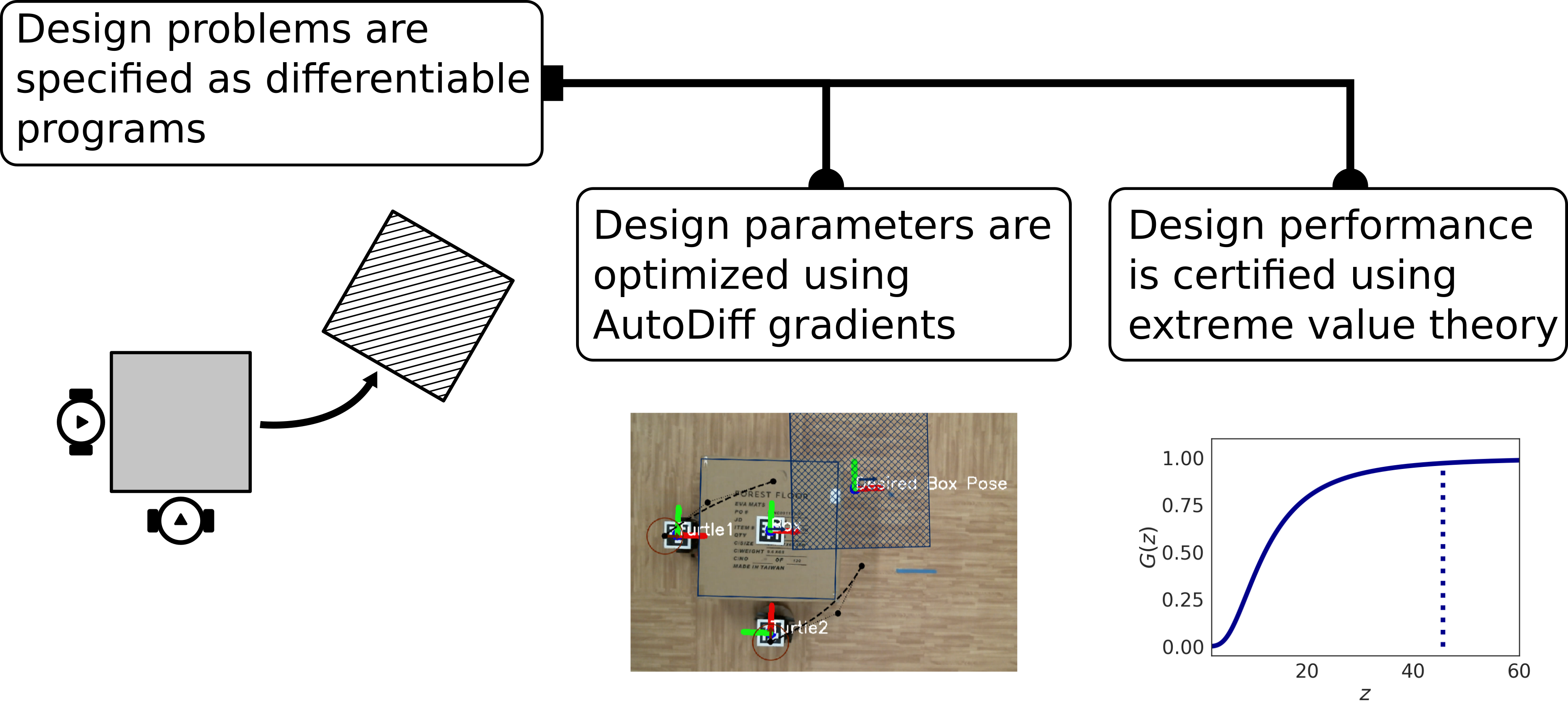}
    \caption{An overview of our framework for robot design optimization and certification. Differentiable programming allows the user to flexibly specify a robot design problem, which can be efficiently optimized using exact gradients and verified using an extreme value statistical analysis.}
    \label{fig:headline}
\end{figure}
Two factors have made it difficult to develop automated design tools for robotics. The first is complexity: most robots are composed of many interacting subsystems. Although some tools may aid in designing certain subsystems (e.g. Simulink for controllers, SolidWorks or CATIA for hardware, custom software for training perception systems), these tools cover only a small part of the overall robotics design problem, which includes sensing, actuation, perception, navigation, control, and decision-making subsystems. In addition to being interconnected, these subsystems often have a large number of parameters that require tuning to achieve good performance (neural network-based perception is an extreme example of this trend). Moreover, since few robotic systems are exactly alike, an effective design tool must allow the user to select an appropriate level of abstraction for the problem at hand. As a result, there is a need for flexible computational tools that can help designers optimize complex robotic systems.

The second difficulty is uncertainty. Robots operate in dynamic environments that cannot be fully specified \textit{a priori}, and nonlinear interactions between the robot and its environment can make this uncertainty difficult to quantify. Nevertheless, we must account for this uncertainty during the design process and ensure that our designs perform robustly. The nature of this uncertainty can vary from problem to problem, reiterating the requirement that an automated design tool must be flexible enough to adapt to different robot design problems.

To be successful, an automated robot design tool must address these two challenges (complexity and uncertainty). In addition, just as mechanical and electrical engineers use automated tools to both \textit{design} and \textit{verify} their designs, a robot design tool must enable its user to both design autonomous systems and certify the robustness of those designs. In this paper, we address these challenges by combining differentiable programming for design optimization with a novel statistical approach to design certification. In short:
\begin{enumerate}
    \item We present a robot design optimization framework that is flexible (using differentiable programming to model complex systems) and robust (avoiding ``brittle'' optima).
    \item We develop a novel statistical approach to certifying a design's robustness to environmental uncertainty.
    \item We validate our approach with experiments in simulation and hardware to show how our methods can be used to solve practical robot design problems.
\end{enumerate}

Our goal is to develop a general-purpose robot design optimization tool that can be applied to a range of robot design problems with multiple subsystems. This goal is in contrast with other approaches that are restricted either to specific applications~\cite{Schulz_robogami,du2016computational,soft_robot_optimization_review,du2021underwater,ma2021diffaqua,zhang_mdo_analysis} or subsystems~\cite{xu_uav_controllers}. To accomplish this goal, we make two novel contributions. The first is algorithmic: our approach builds on recent developments in programming languages (i.e. automatic differentiation) to provide the flexibility to model complex systems while still allowing fast gradient-based optimization. The second concerns certification: to ensure that our optimized designs are robust in the face of uncertainty, we pair design optimization with a novel statistical approach to robustness analysis.

Our experiments show that our methods can (in our first case study) optimize a robotic system with five subsystems and six design variables in under five minutes, achieving an 8.4x performance improvement over the initial design. In our second case study, we optimize a system with three subsystems and 454 design variables in under an hour, achieving a 44\% performance improvement over the initial design. Our use of differentiable programming allows us to complete this optimization 32\% and 20x faster, respectively in each example, compared to approximate gradient methods. Both of these designs are certified using a statistical robustness analysis and successfully deployed in hardware. An open-source implementation of our framework, including repeatable code examples, is available at \url{https://github.com/MIT-REALM/architect}. Our hope is that this prototype implementation will provide the foundation for a fully-featured, easy-to-use design tool for practicing robotics engineers.

\section{Related Work}\label{related_work}

\subsubsection{Design optimization for robotics}

Most existing works on design optimization for robotics focus on a particular application, such as simple walking robots~\cite{Schulz_robogami}, quadrotors~\cite{du2016computational}, and soft robots~\cite{soft_robot_optimization_review,du2021underwater,ma2021diffaqua}. Other works employ optimization to design specific subsystems, such as controllers~\cite{xu_uav_controllers} or motion plans~\cite{schulmanMotionPlanningSequential2014}.
In contrast, the purpose of this work is to develop a general-purpose robot design optimization tool that can be applied not only to a range of robot design problems but also to optimize the design of multiple subsystems simultaneously. This goal is related to that of a large family of multi-disciplinary design optimization (MDO) methods in aerospace engineering~\cite{Martins2013_mdo_survey}. As discussed above, our approach differs from MDO in its use of differentiable programming as a flexible modeling tool and our novel statistical approach to robustness analysis. We review the related work for automatic differentiation and robustness analysis in the next two sections.

\subsubsection{Programming languages for design optimization}

When it comes to managing complexity in a general-purpose design framework, programming languages are a natural tool. They allow users (i.e. programmers) to define precisely which abstractions are appropriate for any given application (e.g. by defining appropriate class hierarchies and function interfaces) without sacrificing generality. To take advantage of this expressivity, we can view engineering designs as programs that define the behavior of the system given suitable choices for design structure and parameters. We can then use automatic differentiation to derive gradients connecting these parameters to the system's behavior and optimize accordingly. This view is inspired by recent work in 3D design optimization~\cite{cascaval2021differentiable}, aircraft design~\cite{sharpe_thesis}, and machine learning~\cite{pytorch,jax2018github}.

In recent years, the robotics community has also developed special-purpose differentiable simulators for robotic systems, particularly those involving rigid body contact dynamics~\cite{heiden2021neuralsim,belubute_peres_lcp_physics,drake,suh2021_bundled_gradients}. These simulators have been used to solve system identification and controller design tasks, but they do not represent a general-purpose framework, as gradients are often derived by hand and the simulators are not expressive enough to model full-stack robotic systems (e.g. with perception and navigation capabilities). We take inspiration from these methods in our case studies, where we implement a simple differentiable contact simulator in our second case study.

\subsubsection{Formal methods for robustness analysis}

Safety and robustness are critical concerns for any robotic system. When it comes to low-level control, there is a rich history of reachability~\cite{althoff_reachability_review} and stability~\cite{chang_neural_lyapunov_control,dawson2021safe} analysis tools that can be used to answer questions of safety and robustness for the control subsystem.
Other works apply reachability analysis at the system level using black-box tools~\cite{fan_dryvr}. This work builds on this history by incorporating formal analysis into a design-optimize-analyze loop to provide rapid feedback on robustness as part of the design process. In particular, we develop a novel statistical method for quantifying the worst-case performance and sensitivity of an optimized design to external perturbations.

\section{Preliminaries and Assumptions}\label{prelim}

Key to the design of robotic systems is the tension between the factors a designer can control and those she cannot. For instance, a designer might be able to choose the locations of sensors and tune controller gains, but she cannot choose the sensor noise or disturbances (e.g. wind) encountered during operation.
Robot design is therefore the process of choosing feasible values for the controllable factors (here referred to as \textit{design parameters}) that achieve good performance despite the influence of uncontrollable factors (\textit{exogenous parameters}). 

Of course, this is a deliberately narrow view of engineering design, since it focuses on parameter optimization and ignores important steps like problem formulation and system architecture selection. Our focus on parameter optimization is intentional, as it allows the designer to focus her creative abilities and engineering judgment on the architecture problem, using computational aids as interactive tools in a larger design process \cite{sharpe_thesis,cascaval2021differentiable}. This focus is common in design optimization (e.g. aircraft design in~\cite{sharpe_thesis} and 3D CAD optimization in~\cite{cascaval2021differentiable}).

\begin{figure}[t!]
    \centering
    \includegraphics[width=\linewidth]{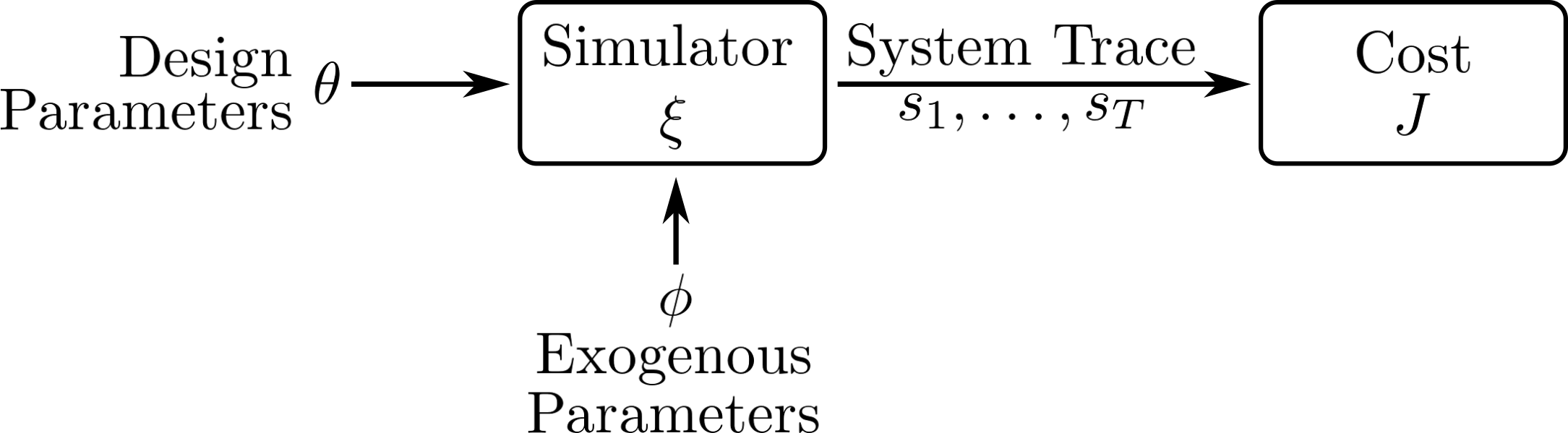}
    \caption{A glass-box model of a generic robotic system. Design optimization involves finding a set of design parameters so that the simulated cost is minimized, while robustness analysis involves quantifying how changes in the exogenous parameters affect the simulated cost.}
    \label{fig:block_diagram}
\end{figure}

To formalize the design optimization problem, we take a high-level view of the robot design problems (shown in Fig.~\ref{fig:block_diagram}), where a design problem has five components:

\subsubsection{Design parameters} The system designer has the ability to tune certain continuous parameters $\theta \in \Theta \subseteq \R^n$; e.g., control gains or the positions of nodes in a sensor network.
\subsubsection{Exogenous parameters} Some factors are beyond the designer's control, such as wind speeds or sensor noise. We model these effects as random variables with some distribution $\phi \sim \Phi$ supported on a subset of $\R^m$. We assume no knowledge of $\Phi$ other than the ability to draw samples i.i.d..
\subsubsection{Simulator} Given particular choices for $\theta$ and $\phi$, the system's state $s \in \mathcal{S}$ evolves in discrete time according to a known simulator $S : \Theta \times \Phi \mapsto \mathcal{S}^T$. This simulator describes the system's behavior over a finite horizon $T$ as a trace of states $s_1, \ldots, s_T$. $S$ should be deterministic; randomness must be ``imported'' via the exogenous parameters.
\subsubsection{Cost} We assume access to a function $J: \mathcal{S}^T \mapsto \mathbb{R}$ mapping system behaviors (i.e. a trace of states) to a scalar performance metric that we seek to minimize.
\subsubsection{Constraints} The choice of design parameters is governed by a set of constraints $c_i : \Theta \mapsto \R$ with index set $i \in \mathcal{I}_c$. Design parameters $\theta$ are feasible if $c_i(\theta) \geq 0 \ \forall i \in \mathcal{I}_c$. Here, we consider constraints as functions of $\theta$ only; we leave the extension to robust constraints involving $\phi$ to future work.

\begin{figure}[tb]
    \centering
    \includegraphics[width=0.8\linewidth]{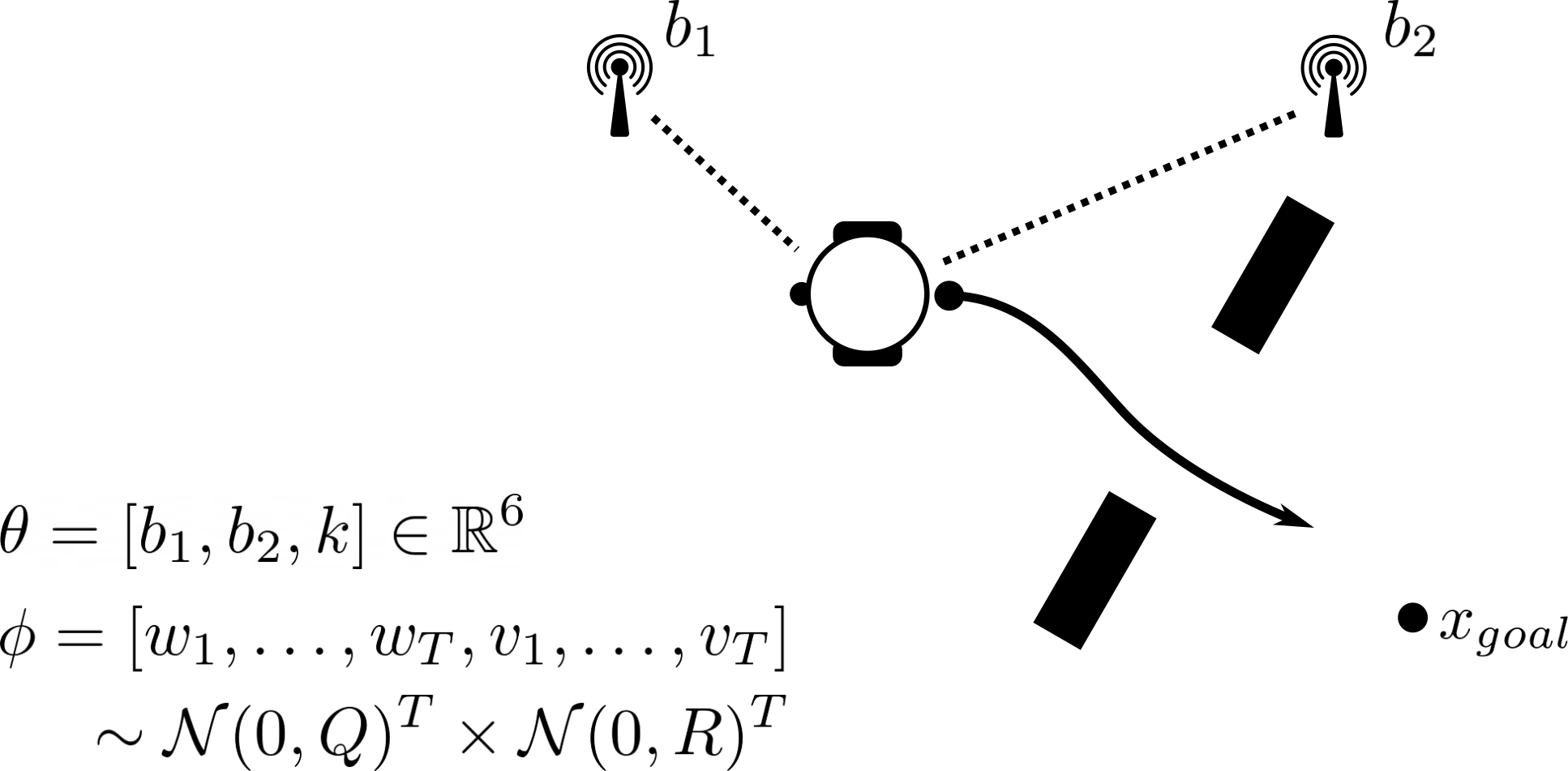}
    \caption{A design optimization problem for an AGV localization and navigation system. The goal is to find placements for two range sensors along with parameters for the navigation system that allow the robot to safely pass through the narrow doorway.}
    \label{fig:agv_example}
\end{figure}

We can make this discussion concrete with an example: consider the autonomous ground vehicle (AGV) design problem illustrated in Fig.~\ref{fig:agv_example}. In this problem, our goal is to design a localization and navigation system that will allow the AGV to safely navigate between two obstacles. The AGV can estimate its position using an extended Kalman filter (EKF) with noisy measurements of its range from two nearby beacons and its heading from an IMU. The robot uses this estimate with a navigation function~\cite{aiama} and feedback controller to track a collision-free path between the obstacles.

In this problem, the design parameters $\theta$ include the $(x, y)$ locations of the two range beacons $b_1, b_2 \in \R^2$ and the feedback controller gains $k \in \R^2$. The exogenous parameters $\phi$ are the actuation and sensor noises at each timestep $w_t \in \R^3$ and $v_t \in \R^3$, drawn i.i.d. from Gaussian distributions $\mathcal{N}(0, Q)$ and $\mathcal{N}(0, R)$, respectively, as well as the initial state (also Gaussian). The simulator $\xi$ integrates the AGV's dynamics using a fixed timestep, updating the EKF and evaluating the navigation controller at each step. The cost function $J$ assigns a penalty to collisions with the environment, estimation errors, and deviations from the goal location. We will return to this example in more detail in Section~\ref{case1}; first, we discuss our approach to design optimization and robustness analysis in Sections~\ref{optimization} and~\ref{analysis}, respectively.

\section{Design Optimization}\label{optimization}

Given the notation from Section~\ref{prelim}, we can formally pose the robot design optimization problem. In formulating the optimization objective, it is important to consider the variance introduced by the exogenous parameters $\phi$. Simply minimizing the expected value of the cost $\expectation_{\phi \sim \Phi} \big[ J\circ S\pn{\theta, \phi} \big]$ (where $\circ$ denotes composition) can lead to myopic behavior where exceptional performance for some values of $\phi$ compensates for poor performance on other values; this is related to the phenomenon of ``reward hacking'' in reinforcement learning~\cite{amodei2016_ai_safety}.

Ideally, we would like our designs to be robust to variations in exogenous parameters: changing $\phi$ should not cause the performance to change much. We can include this requirement as a heuristic by penalizing the variance of $J$. Intuitively, this heuristic ``smooths'' the cost function with respect to the exogenous parameters: regions of high variance (containing sharp local minima) are penalized, while regions of low variance are rewarded. We return to justify this connection to robustness in Section~\ref{connection_to_robustness}. This heuristic leads us to the \textit{variance-regularized robust design optimization problem}:
\begin{subequations}\label{design_optimization_nlp_generic}
\begin{align}
    \min_{\theta \in \Theta} &\quad \expectation_{\phi \sim \Phi} \Big[ J\circ S\pn{\theta, \phi} \Big] + \lambda \rm{Var}_{\phi\sim\Phi}\Big[ J\circ S\pn{\theta, \phi} \Big] \label{design_optimization_objective_generic} \\
    \text{s.t.} &\quad c_i(\theta) \geq 0 \quad \forall i \in \mathcal{I}_c \label{design_optimization_constraints_generic}
\end{align}
\end{subequations}
Practically, we replace the expectation and variance with unbiased estimates over $N$ samples $\phi_i \sim \Phi, i=1,\ldots,N$.
\begin{subequations}\label{design_optimization_nlp}
\begin{align}
    \min_{\theta \in \Theta} &\quad \frac{1}{N}\sum_{i=1}^{N} \Big[J\circ S\pn{\theta, \phi_i}\Big] \label{design_optimization_objective} \\
    &+ \lambda \left[ \frac{\sum_{i=1}^N \pn{J\circ S\pn{\theta, \phi_i}}^2}{N-1} - \frac{\pn{\sum_{i=1}^N J\circ S\pn{\theta, \phi_i}}^2}{(N-1)N} \right] \nonumber \\
    \text{s.t.} &\quad c_i(\theta) \geq 0 \quad \forall i \in \mathcal{I}_c \label{design_optimization_constraints}
\end{align}
\end{subequations}

Of course, these Monte-Carlo estimators will require multiple evaluations of $J\circ S$ to evaluate~\eqref{design_optimization_objective}. Since $S$ might itself be expensive to evaluate, approximating the gradients of \eqref{design_optimization_objective} and \eqref{design_optimization_constraints} using finite differences will impose a large computational cost ($2nN$ additional evaluations of $J\circ S$ and $c_i$ at each step). Instead, we can turn to automatic differentiation (AD) to directly compute these gradients with respect to $\theta$, which we can use with any off-the-shelf gradient-based optimization engine. The precise choice of optimization algorithm is driven by the constraints and is not central to our framework. If the constraints are hyper-rectangle bounds on $\theta$, then algorithms like L-BFGS-B may be used, but if the constraints are more complex then sequential quadratic programming or interior-point methods may be used. Our implementation provides an interface to a range of optimization back-ends through SciPy~\cite{2020SciPy-NMeth}, and we plan to add support for hybrid methods combining local gradient descent with gradient-free population methods in a future work.

In this framework, the user need only implement the simulator and cost function for their specific problem using a differentiable programming framework like the JAX library for Python~\cite{jax2018github}, and this implementation can be used automatically for efficient gradient-based optimization. By implementing a library of additional building blocks in this AD paradigm (e.g. estimation algorithms like the EKF), we can provide an AD-based design optimization tool that strikes a productive balance between flexibility and ease of use. In the supplementary materials, we provide a prototype implementation of this tool, containing some of these AD building blocks. In future work, we hope to further expand this library to include more common robotics algorithms.

\section{Design Certification via Robustness Analysis}\label{analysis}

Once we have found an optimal choice of design parameters, we need to verify that the design will be robust to uncertainty in the exogenous parameters. Similarly to 3D CAD and FEA packages for mechanical engineers, a successful design tool not only helps an engineer refine her design (i.e. using the design optimization framework in Section~\ref{optimization}) but also helps her analyze and predict its performance. To certify the performance of an optimized design, we are interested in two distinct questions. First, what is the maximum cost we can expect given variation in the exogenous parameters? Second, how sensitive is the cost to external disturbances: by how much can a change in the exogenous parameters increase the cost?

Answering these questions is difficult because we must extrapolate from a finite number of simulations to predict worst-case performance. To address this difficulty, we develop a probabilistic approach based on extreme value theory in statistics~\cite{sridhar2021improving,wood_zhang_1996,coles_2001}. We begin by stating a relevant result:

\begin{theorem}[Extremal Types Theorem; 3.1.1 in ~\cite{coles_2001}]\label{extreme_value_thm}
Let $X_1, \ldots, X_N$ be random variables drawn i.i.d. from an unknown distribution and $M_N = \max_i \{ X_i \}$ be the sample maximum. If there exist sequences of normalizing constants $\{a_N > 0\}$ and $b_N$ such that the limiting distribution of $(M_N - b_N)/a_N$ as $N\to\infty$ is non-degenerate, then
\begin{equation}
    \lim_{N\to\infty} \Pr\left[(M_N - b_N) / a_N \leq z \right] = G(z)
\end{equation}
where $G(z)$ is a Generalized Extreme Value Distribution (GEVD) with location $\mu$, scale $\sigma$, and shape $\xi$,
\begin{equation}
    G(z) = \exp \left\{ -\left[ 1 + \xi\pn{\frac{z - \mu}{\sigma}} \right]^{-1/\xi} \right\},
\end{equation}
supported on $\{z : 1 + \xi(z - \mu)/\sigma > 0\}$.
\end{theorem}

In the special case $\xi = 0$, this distribution has a slightly different form (known as a Gumbel distribution), but the result holds. In practice, $a_n$ and $b_n$ are not estimated directly (this merely changes the fit values of $\mu$ and $\sigma$) and the GEVD is fit directly to $M_N$ by either minimizing the log likelihood~\cite{coles_2001} or estimating the posterior distribution of $(\mu, \sigma, \xi)$ using Markov Chain Monte Carlo sampling~\cite{salvatier_wiecki_fonnesbeck_2016}. A useful feature of the GEVD is that if our data suggest that $\xi < 0$, then the support of $G(z)$ is bounded above and we can estimate an upper bound on the maximum $M_\infty$. If $\xi \geq 0$, then we cannot estimate a strict upper bound, but we can provide for a confidence interval for $M_\infty$ instead. In the following sections, we apply this theorem to analyze the robustness of an optimized design.

\subsection{Estimating the worst-case performance}

Our first robustness question concerns the worst-case performance of our design: given variation in $\phi$, what is the maximum cost\footnote{Any function of the simulation trace can be substituted for cost without changing the framework.} we can expect for our choice of design parameters $\theta$? Our insight is that the variation $\phi \sim \Phi$ induces an (unknown) distribution in $J \circ S (\theta, \phi)$, so $J \circ S (\theta, \phi)$ a random variable to which the extremal types theorem applies. Algorithm~\ref{alg:worst_case_cost} provides a means for estimating the maximum of $J \circ S (\theta, \phi)$ by fitting a GEVD to observed maximums $M_N$. Generally speaking, the block size $N$ and sample size $M$ should be chosen to be as large as computationally feasible to reduce the variance of the GEVD estimate~\cite{coles_2001}.

\begin{algorithm}
\caption{An algorithm for estimating the parameters of a GEVD governing the expected maximum cost $J \circ S$}\label{alg:worst_case_cost}
    \begin{algorithmic}
        \Require Block size $N > 0$ and sample size $M > 0$
        \State $X_j^i \gets J\circ S(\theta, \phi_{ij})$; with $\phi_{ij}\sim\Phi$, $1\leq j\leq N$, $1\leq i\leq M$
        \State $M_N^i \gets \max\{X_1^i, \ldots, X_N^i\}$ for $i=1,\ldots,M$
        \State $(\mu, \sigma, \xi) \gets$ posterior GEVD estimate given $\{M_N^i\}$
    \end{algorithmic}
\end{algorithm}

In practice, we use the automatic parallelization features of JAX to efficiently compute $X_j^i$ and obtain the posterior distribution of $\mu$, $\sigma$, and $\xi$ using Markov Chain Monte Carlo sampling with the PyMC3 library~\cite{salvatier_wiecki_fonnesbeck_2016}. From this posterior distribution, we take the $97\%$ confidence level for each parameter $(\mu^*, \sigma^*, \xi^*)$. If $\xi^* < 0$, we have confidence that the corresponding GEVD has bounded support on the right and estimate the maximum cost $J_{max} \leq \mu - \sigma/\xi$. Otherwise, we can estimate the 97\% confidence level for $J_{max}$ using the GEVD described by $(\mu^*, \sigma^*, \xi^*)$.

\subsection{Estimating sensitivity}

In addition to the expected worst-case performance, it is also useful to know the sensitivity of that performance. That is, if the design performs well in one situation (i.e. for some value of $\phi$), then how much can we expect its performance to degrade if $\phi$ changes? Formally, we define the sensitivity $L$ as the least constant such that for any two $\phi_1, \phi_2 \sim \Phi$, $$|J\circ S(\theta, \phi_1) - J\circ S(\theta, \phi_2)| \leq L ||\phi_1 - \phi_2||$$ 
If $J \circ S$ is Lipschitz then $L$ will be finite and equal the Lipschitz constant of $J \circ S$, but we do not require this assumption; if $J \circ S$ is not Lipschitz, then we can estimate a high-confidence upper bound on $L$.

In both cases, we can exploit the fact that $L$ is an extreme value of the slope $|J\circ S(\theta, \phi_1) - J\circ S(\theta, \phi_2)| / ||\phi_1 - \phi_2||$ and apply the extremal types theorem. Let $X = ||J\circ S(\theta, \phi_1) - J\circ S(\theta, \phi_2)|| / ||\phi_1 - \phi_2||$ be a random variable with $\phi_1, \phi_2 \sim \Phi$. The distribution of $X$ is unknown, but the extremal types theorem lets us characterize the sample maximum $L_N = \max\{X_1, \ldots, X_N\}$ using a GEVD. Algorithm~\ref{alg:sensitivity} provides our method for fitting this distribution, and a concrete Python implementation is provided in the supplementary materials. This approach is similar to that in~\cite{sridhar2021improving,knuth_chou_2021} but removes the assumption that $L$ is bounded by fitting a GEVD instead of a reverse Weibull distribution, allowing our approach to apply when $J\circ S$ is not Lipschitz.
\begin{algorithm}
\caption{An algorithm for estimating the parameters of a GEVD governing the sensitivity of $J \circ S$}\label{alg:sensitivity}
    \begin{algorithmic}
        \Require Block size $N > 0$ and sample size $M > 0$
        \State $X_j^i \gets |J\circ S(\theta, \phi_{ij,1}) - J\circ S(\theta, \phi_{ij,2})| / ||\phi_{ij,1} - \phi_{ij,2}||$, \hspace{2cm} \phantom{xxxxxxx} with $\phi_{ij,1}, \phi_{ij,2} \sim\Phi$, $j=1,\ldots,N$, $i=1,\ldots,M$
        \State $L_N^i \gets \max\{X_1^i, \ldots, X_N^i\}$ for $i=1,\ldots,M$
        \State $(\mu, \sigma, \xi) \gets$ posterior GEVD estimate given $\{L_N^i\}$
    \end{algorithmic}
\end{algorithm}

Algorithm~\ref{alg:sensitivity} is similar to Algorithm~\ref{alg:worst_case_cost}, but the interpretation of the results differs in that the fit parameters from Algorithm~\ref{alg:sensitivity} allow us to understand the sensitivity of a design. In particular, if the 97\% confidence level for the shape parameter $\xi^*$ is negative, then $J\circ S$ is likely Lipschitz continuous with Lipschitz constant $L \leq \mu - \frac{\sigma}{\xi}$. If $\xi > 0$, then $J\circ S$ is likely not Lipschitz but we can estimate the 97\% confidence level for $L$. As a result, this statistical approach allows us to avoid making prior assumptions about the continuity of our system.


\subsection{Connections to Design Optimization}\label{connection_to_robustness}

Here, we will attempt to justify the variance regularization heuristic introduced in Section~\ref{optimization} with reference to the worst-case performance $J_{max}$ and sensitivity $L$ computed by Algorithms~\ref{alg:worst_case_cost} and~\ref{alg:sensitivity}. First, let's examine the connection with expected worst-case performance $J_{max}$. If we take the probability of observing a cost $J = J\circ S(\theta, \phi)$ within $\alpha$ of $J_{max}$ ($0 < \alpha < 1$) and apply Cantelli's inequality~\cite{boucheron_lugosi_massart_2016}, we see that
\begin{align*}
    \rm{Pr}_{\phi \sim \Phi}(J \geq \alpha J_{max}) &\leq \frac{\rm{Var}_{\phi \sim \Phi}[J]}{\rm{Var}_{\phi \sim \Phi}[J] + (\alpha J_{max} - \expectation_{\phi \sim \Phi}[J])^2}
\end{align*}
Minimizing $\rm{Var}_{\phi \sim \Phi}[J]$ in addition to $\expectation_{\phi \sim \Phi}[J]$ will correlate with decreasing this upper bound. As a result, we expect variance regularization to correlate with decreased probability of encountering near-worst-case performance.

We can also justify the connection between variance regularization and reducing sensitivity $L$ by looking at the special case where $J \circ S$ is Lipschitz and the elements of $\phi$ are independent. The Bobkov-Houdr\'e variance bound for Lipschitz functions~\cite{bobkov} holds that $\rm{Var}_{\phi \sim \Phi}[J] \leq L^2 \sigma_{\Sigma\phi}^2$,
where $\sigma_{\Sigma\phi}^2$ is the variance of the sum of elements in $\phi$. This bound does not explicitly show that minimizing $\rm{Var}_{\phi \sim \Phi}[J]$ decreases $L$, but it suggests a correlation that we hope to revisit in future work.

\section{Experimental Results}\label{cases}

So far, we have developed the theoretical and algorithmic basis for our robot design framework. It remains for us to empirically answer two questions: first, is our framework useful for solving practical robot design problems? Second, is our statistical method for robustness analysis sound?

In this section, we answer these questions through the lens of two case studies. The first involves finding optimal sensor placements for robot navigation, and the second involves optimizing a pushing strategy for multi-agent manipulation. We demonstrate the success of our optimization and robustness analysis framework on each example, and we provide results from hardware testing in both cases. Next, we include an ablation study justifying our use of automatic differentiation and variance regularization. We conclude by verifying the soundness of our statistical robustness analysis.

\subsection{Case study: optimal sensor placement for navigation}\label{case1}

First, we return to the AGV localization and navigation example introduced in Fig.~\ref{fig:agv_example}. This design problem requires finding an optimal placement for two ranging beacons to minimize estimation error and allow the robot to safely navigate between two obstacles. Range measurements from these beacons are integrated with IMU data via an EKF, and the resulting state estimate is used as input to a navigation function and tracking feedback controller to guide the robot to its goal. This design problem has two important features. First, it involves interactions between multiple subsystems: the output from the EKF is used by the navigation function, which feeds input to the controller, which in turn influences future EKF predictions. Second, the effect of uncertainty on the robot's performance is relatively strong.

The design parameters are the $(x, y)$ locations of two range beacons and two feedback controller gains (6 total design parameters). The exogenous parameters include uncertainty in the robot's initial state along with actuation and sensing noise at each of $T$ timesteps ($3 + 6T$ total exogenous parameters). The cost function has three components: one penalizing large estimation errors, one penalizing deviations from the goal, and one penalizing collisions with the environment. A formal definition of the design and exogenous parameters, simulator, cost, and constraints is given in Table~\ref{tab:agv_design_problem} in the appendix. We also include code in the supplementary materials for defining this design problem in our framework and running our design optimization and sensitivity analysis methods. The simulator and cost functions are implemented in Python using the JAX framework for automatic differentiation.

\begin{figure}[tb]
    \centering
    \includegraphics[width=0.8\linewidth]{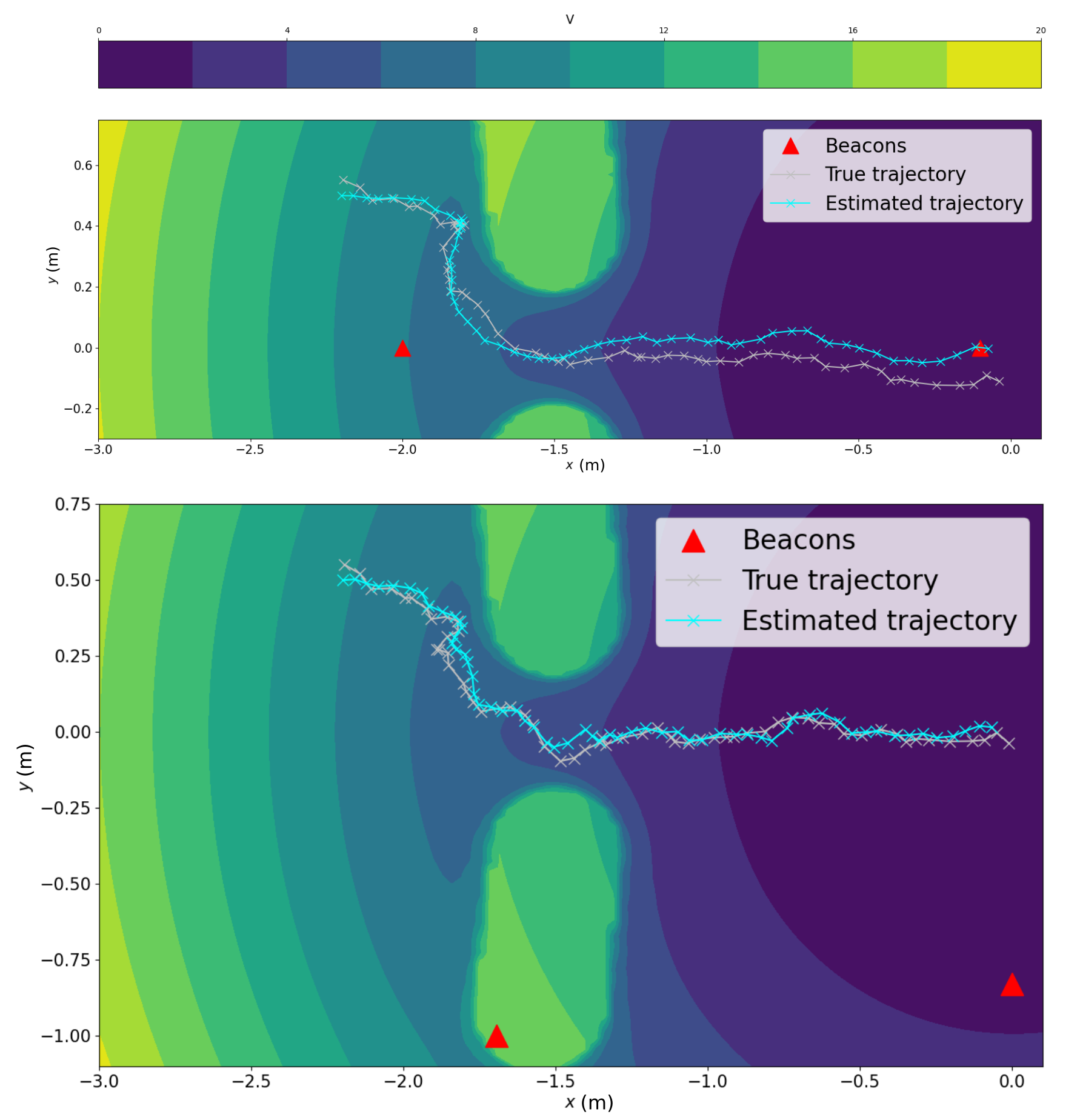}
    \caption{Simulated trajectories for the initial (top) and optimized (bottom) AGV designs. Color indicates the value of the navigation function. Beacon positions are bounded within the area shown.}
    \label{fig:agv_representative_trajectories}
\end{figure}

Fig.~\ref{fig:agv_representative_trajectories} compares simulated trajectories for the initial and optimized beacon placements and feedback gains, clearly showing the impact of design optimization. Initially, poor beacon placement causes the robot to accumulate estimation error and drift away from its goal. The optimized design moves the beacons off to the side to eliminate this drift. Optimization ($N=512$, $\lambda=0.1$, L-BFGS-B back-end) took 3 minutes \SI{34}{s} on a laptop computer (\SI{8}{GB} RAM, \SI{1.8}{GHz} 8-core processor).

\begin{figure}[tb]
    \centering
    \begin{subfigure}[t]{0.5\linewidth}
        \centering
        \includegraphics[width=\linewidth]{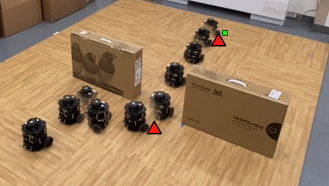}
    \end{subfigure}%
    \begin{subfigure}[t]{0.5\linewidth}
        \centering
        \includegraphics[width=\linewidth]{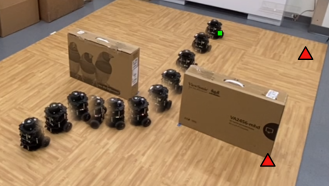}
    \end{subfigure}
    \caption{Hardware performance of initial (left) and optimized (right) AGV designs. Square (green) shows the goal; triangles (red) show beacon locations. The optimized design eliminates drift relative to goal.}
    \label{fig:agv_hw}
\end{figure}

\begin{figure}[tb]
    \centering
    \includegraphics[width=\linewidth]{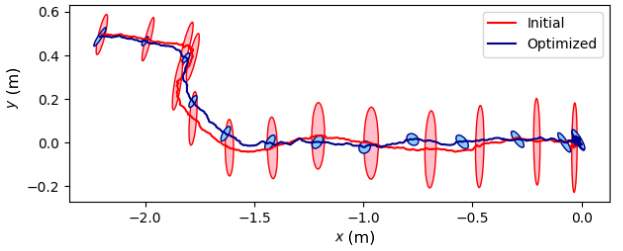}
    \caption{Hardware results for EKF state estimates and posterior error covariance $3\sigma$ ellipse for initial and optimized designs.}
    \label{fig:agv_hw_cov}
\end{figure}

We tested the initial and optimized design in hardware using the Turtlebot 3 platform. To emulate range beacon measurements in our lab, odometry and laser scan data were fused into a full state estimate from which range measurements were derived (the full state estimate was hidden from the robot, which only received the emulated range measurements). The control frequency was increased from \SI{2}{Hz} in simulation to \SI{10}{Hz} in hardware, and the obstacles were recreated in our laboratory. The hardware results, shown in Figs.~\ref{fig:agv_hw} and~\ref{fig:agv_hw_cov}, confirm our simulation results: the initial design suffers from drift and ends approximately \SI{10}{cm} from its target position, while the optimized design does not drift and ends within \SI{5}{cm} of the goal. This difference can be seen most clearly in the posterior error covariance from the EKF;  Fig.~\ref{fig:agv_hw_cov} shows how the optimized design greatly reduces uncertainty in the state estimate compared to the initial design. No parameter estimation or tuning was required.

\begin{figure}[t]
    \centering
    \includegraphics[width=0.9\linewidth]{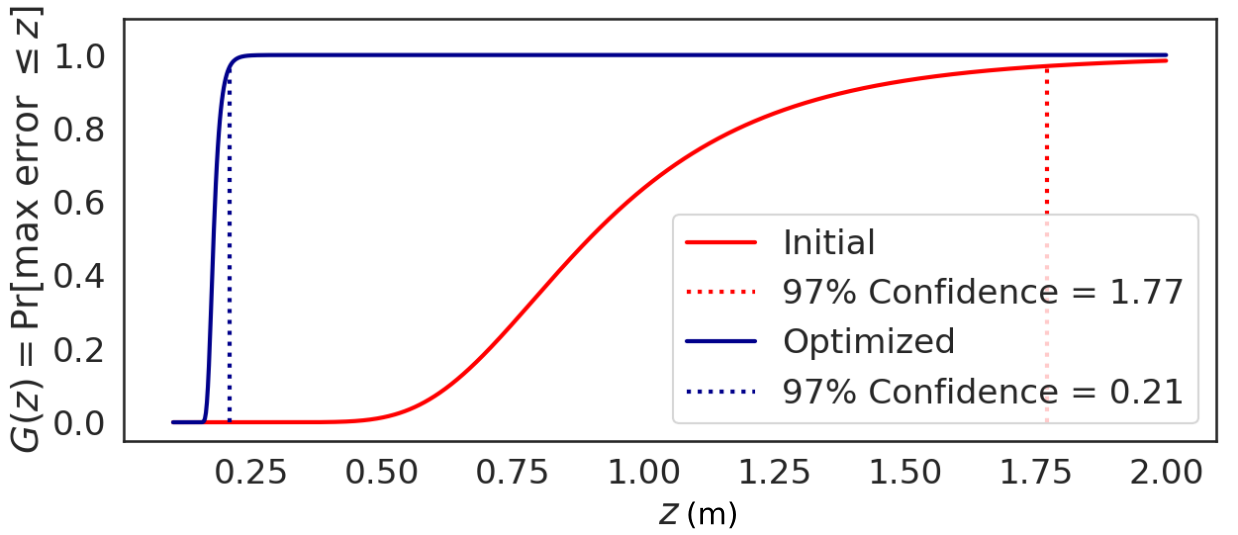}
    \caption{GEVD CDF fit using Algorithm~\ref{alg:worst_case_cost} for the maximum absolute estimation error in the $xy$-plane in both the initial and optimized designs, with 97\% confidence levels.}
    \label{fig:agv_gevd}
\end{figure}

Finally, we apply the robustness analysis from Section~\ref{analysis} to certify the maximum absolute estimation error $\norm{x_t - \hat{x}_t}$ in the optimized design (in meters, projected into the $xy$ plane). Note that this error is different from the cost used during optimization, but we can still apply Algorithm~\ref{alg:worst_case_cost} simply by changing the cost function for the duration of the analysis. Using block size $N = 1000$ and sample size $M = 1000$, we fit a GEVD using Algorithm~\ref{alg:worst_case_cost} to the maximum estimation error for both the initial and optimized designs. These distributions are shown in Fig.~\ref{fig:agv_gevd}; the optimized design significantly reduces the expected maximum estimation error. We observe that the 97\% confidence level for the shape parameter $\xi = 0.059$ is positive, so we cannot conclude that the worst-case estimation error is bounded, but we can derive a high-confidence bound of \SI{0.21}{m} for our optimized design.

\subsection{Case study: collaborative multi-robot manipulation}\label{case2}

\begin{figure}[b!]
    \centering
    \includegraphics[width=0.5\linewidth]{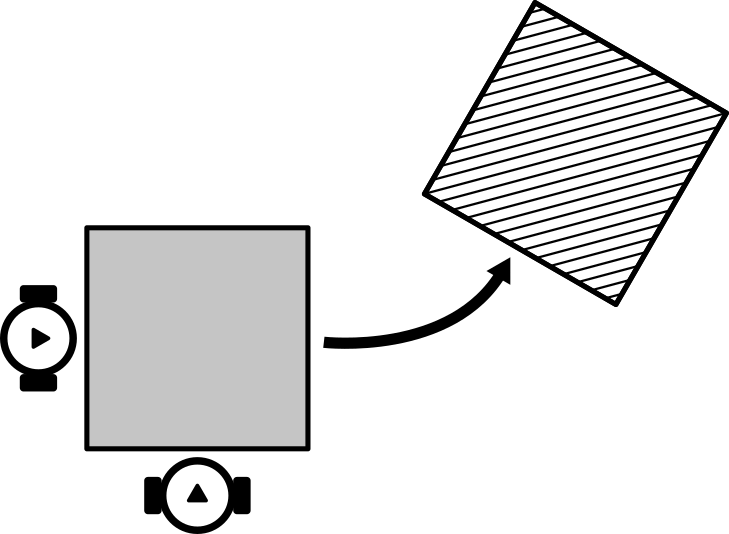}
    \caption{Multi-agent manipulation design optimization problem. The goal is to find parameters for robot controllers and a neural network planner that push the box from an initial position (solid) to a desired position (striped).}
    \label{fig:mam_example}
\end{figure}

Our second example involves finding a control strategy for multi-agent collaborative manipulation. In this setting, two ground robots must collaborate to push a box from its current location to a target pose (as in Fig.~\ref{fig:mam_example}). Given the desired box pose and the current location of each robot, a neural network plans a trajectory for each robot, which the robots then track using a feedback controller ($\theta$ includes both the neural network parameters and the tracking controller gains, with a total of 454 design parameters). The exogenous parameters include the coefficient of friction for each contact pair, the mass of the box, the desired pose of the box, and the initial pose for each robot (a total of 13 exogenous parameters; we vary the desired box pose and initial robot poses to prevent over-fitting during optimization). The cost function is simply the squared error between the desired box pose (including position and orientation) and its true final pose after a \SI{4}{s} simulation. A full definition of this design problem and contact dynamics model is included in Table~\ref{tab:mam_design_problem} in the appendix. We implement the contact dynamics simulator, trajectory planning neural network, and path tracking controller in Python using JAX.

\begin{figure}[t]
    \centering
    \includegraphics[width=0.8\linewidth]{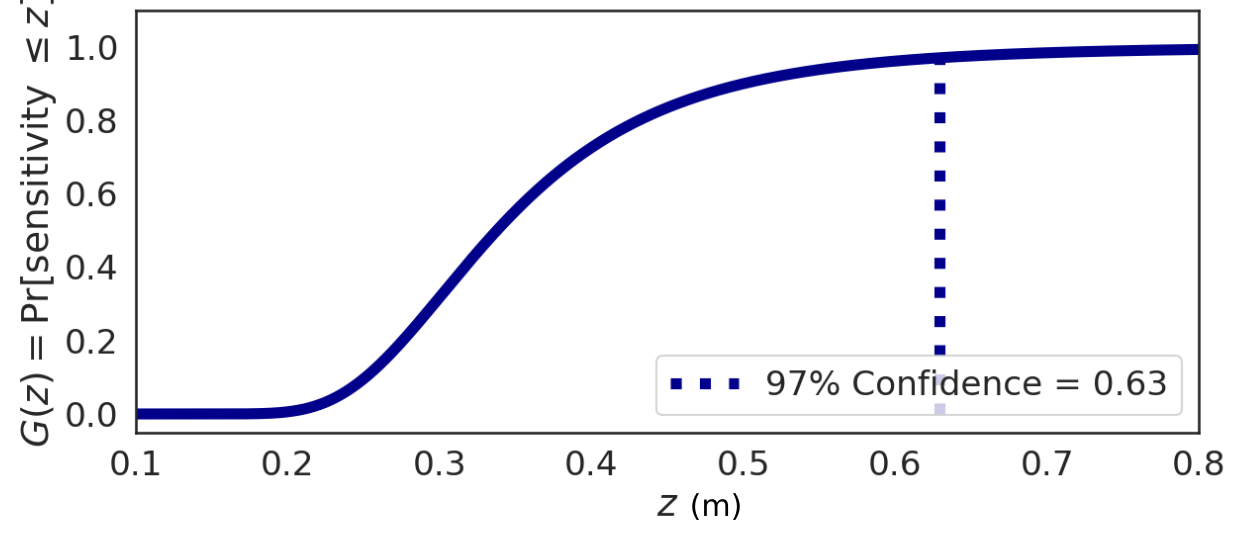}
    \caption{GEVD CDF fit using Algorithm~\ref{alg:sensitivity} for the maximum sensitivity of the optimized collaborative manipulation strategy to variation in friction coefficient. $z$ has units of meters per unit change in friction coefficient.}
    \label{fig:mam_gevd}
\end{figure}

\begin{figure*}[tb!]
    \centering
    \includegraphics[width=0.9\linewidth]{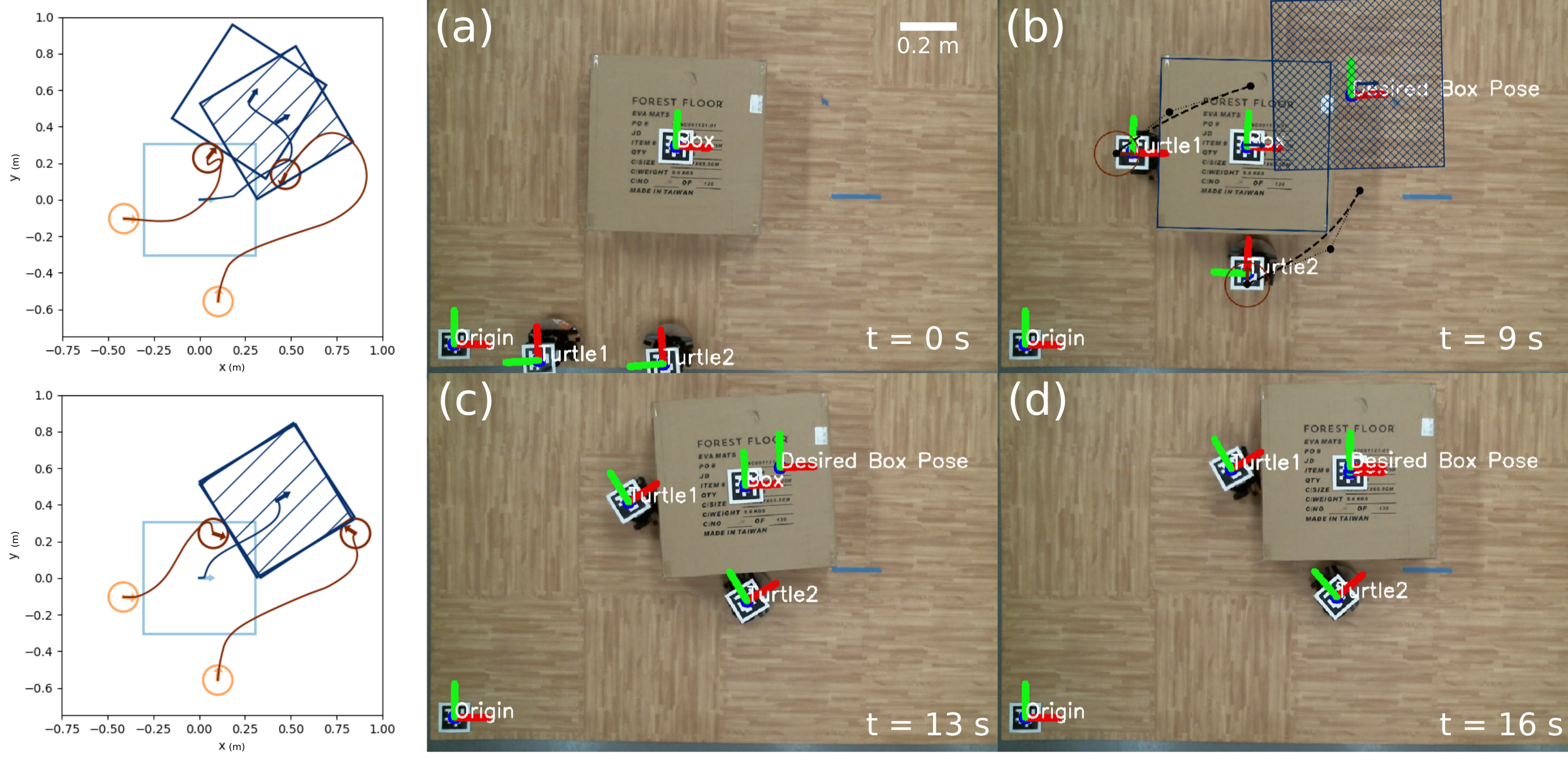}
    \caption{Left: Initial (top) and optimized (bottom) manipulation strategies in simulation (light/dark colors indicate initial/final positions, stripes indicate desired position). Right: Optimized manipulation strategy deployed in hardware (video included in the supplementary materials). (a) The robots first move to positions around the box. (b) Using the optimized neural network, the robots plan a cubic spline trajectory pushing the box to its desired location. (c-d) The robots execute the plan by tracking that trajectory.}
    \label{fig:mam_hw}
\end{figure*}


Compared to the design problem in our first case study, this system has a simpler architecture (fewer subsystems) but more complicated dynamics and a much higher-dimensional design space. This example also showcases a different interpretation of the exogenous parameters: instead of representing true sources of randomness, these parameters represent quantities that are simply unknown at design-time. For example, the target position for the box is not random in the same way as sensor noise in the previous example, but since we cannot choose this value at design-time it must be included in $\phi$. As a result, minimizing the expected cost with respect to variation in $\phi$ yields a solution that achieves good performance for many different target poses, enabling the user to select one at run-time and be confident that the design will perform well.

To solve this design problem, the neural network parameters are initialized i.i.d. according to a Gaussian distribution, and the tracking controller gains are set to nominal values. We then optimize the parameters using $N = 512$, $\lambda = 0.1$, and L-BFGS-B back-end. This optimization took 45 minutes \SI{32}{s} on a laptop computer (\SI{8}{GB} of RAM and a \SI{1.8}{GHz} 8-core processor). Fig.~\ref{fig:mam_hw} shows a comparison between the initial and optimized strategies, and Fig.~\ref{fig:mam_more} in the appendix shows additional examples of the optimized behavior. The target pose is drawn uniformly $[x, y, \theta] \in[0, 0.5]^2 \times [-\pi/4, \pi/4]$, and the optimized design achieves a mean squared error of $0.0964$.

We tested the optimized design in hardware, again using the Turtlebot 3 platform. An overhead camera and AprilTag~\cite{olson2011tags} markers were used to obtain the location of the box and each robot. At execution, each robot first moves to a designated starting location near the box, plans a trajectory using the neural network policy, and tracks that trajectory at \SI{100}{Hz} until the box reaches its desired location or a time limit is reached. Results from this hardware experiment are shown in Fig.~\ref{fig:mam_hw}, and a video is included in the supplementary materials. Again, no parameter tuning or estimation was needed.

After successfully testing the optimized design in the laboratory, it is natural to ask how its performance might change as conditions (particularly the coefficients of friction) change. Using $M = 500$ blocks of size $N=1000$ each, we use Algorithm~\ref{alg:sensitivity} to fit a GEVD for the sensitivity constant $L$ with respect to the coefficients of friction between each contact pair. We do this by allowing these coefficients to vary and freezing other elements of $\phi$ at nominal values (box mass \SI{1}{kg} and target pose $[0.3, 0.3, 0.3]$). The fit distribution is shown in Fig.~\ref{fig:mam_gevd}. The 97\% confidence level for the shape parameter is $\xi = 0.118 > 0$, so we cannot conclude that the performance of our design is Lipschitz with respect to the friction coefficients, but we can estimate the 97\% confidence level for $L$ as $0.63$.

\subsection{Design optimization ablation study}\label{ablation}

\begin{figure*}[t]
    \centering
    \begin{subfigure}[t]{0.25\linewidth}
        \centering
        \includegraphics[width=\linewidth]{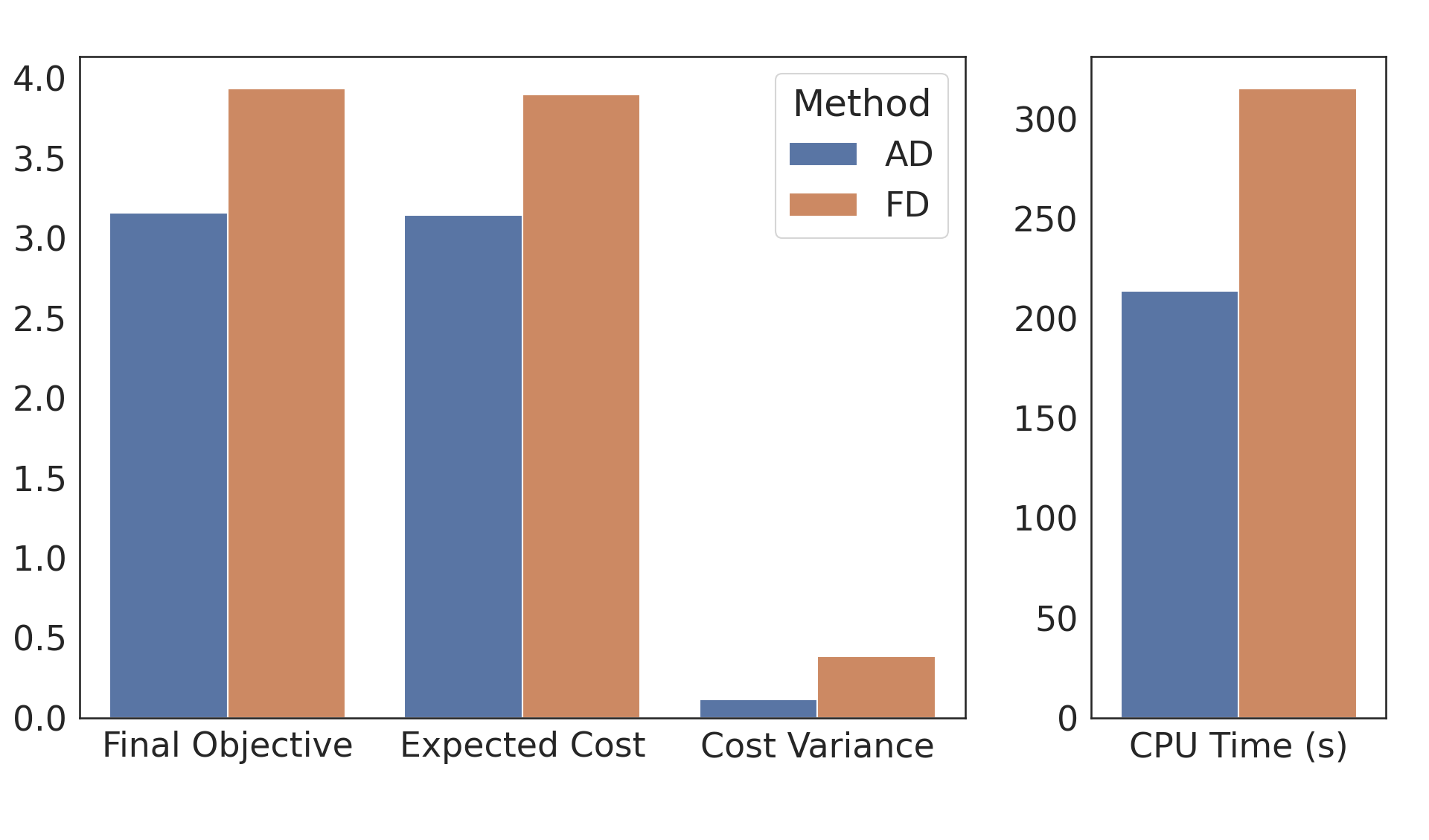}
        \caption{AD vs. FD; sensor placement}
    \end{subfigure}%
    \begin{subfigure}[t]{0.25\linewidth}
        \centering
        \includegraphics[width=\linewidth]{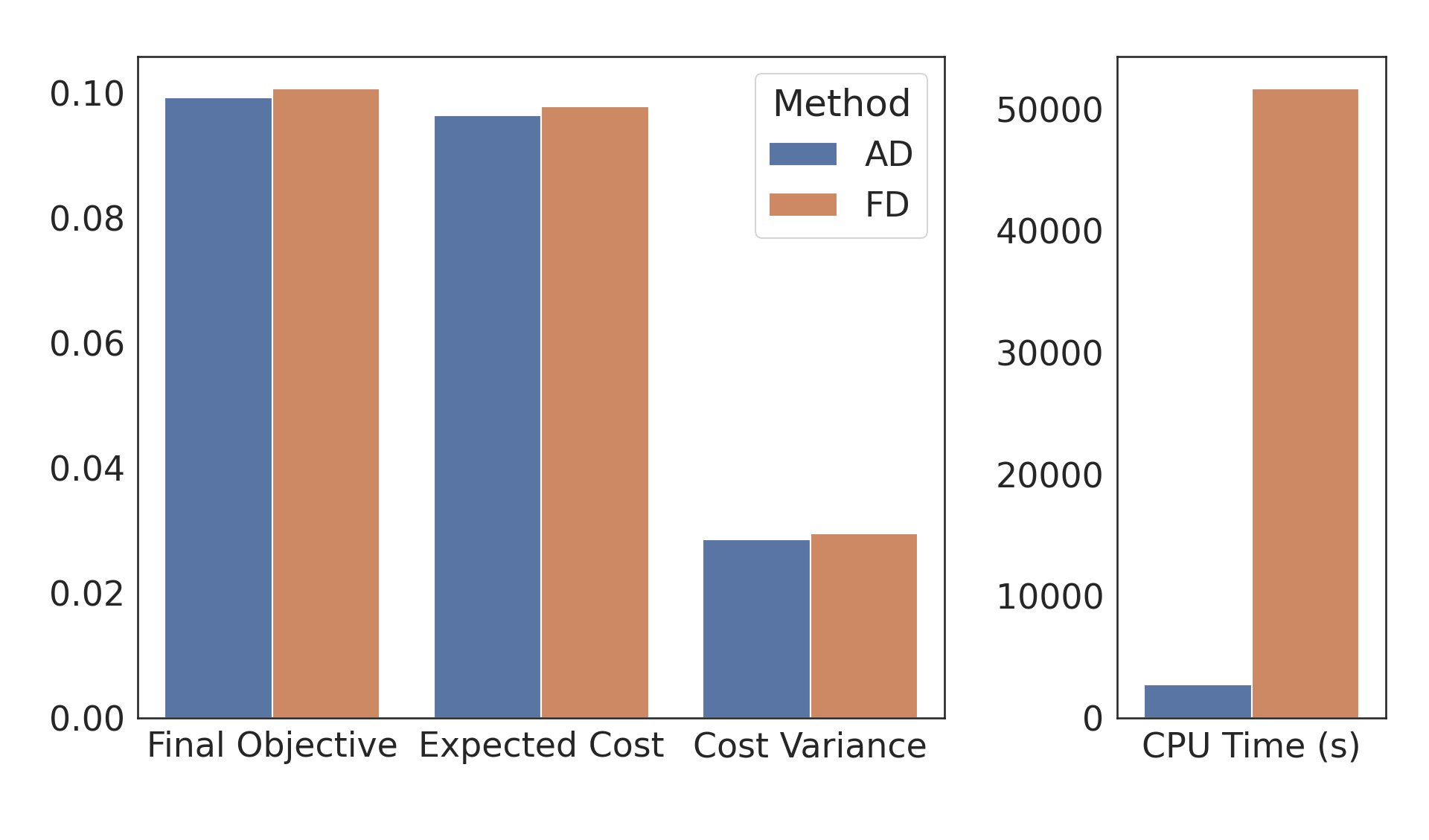}
        \caption{AD vs. FD; manipulation}
    \end{subfigure}%
    \begin{subfigure}[t]{0.25\linewidth}
        \centering
        \includegraphics[width=\linewidth]{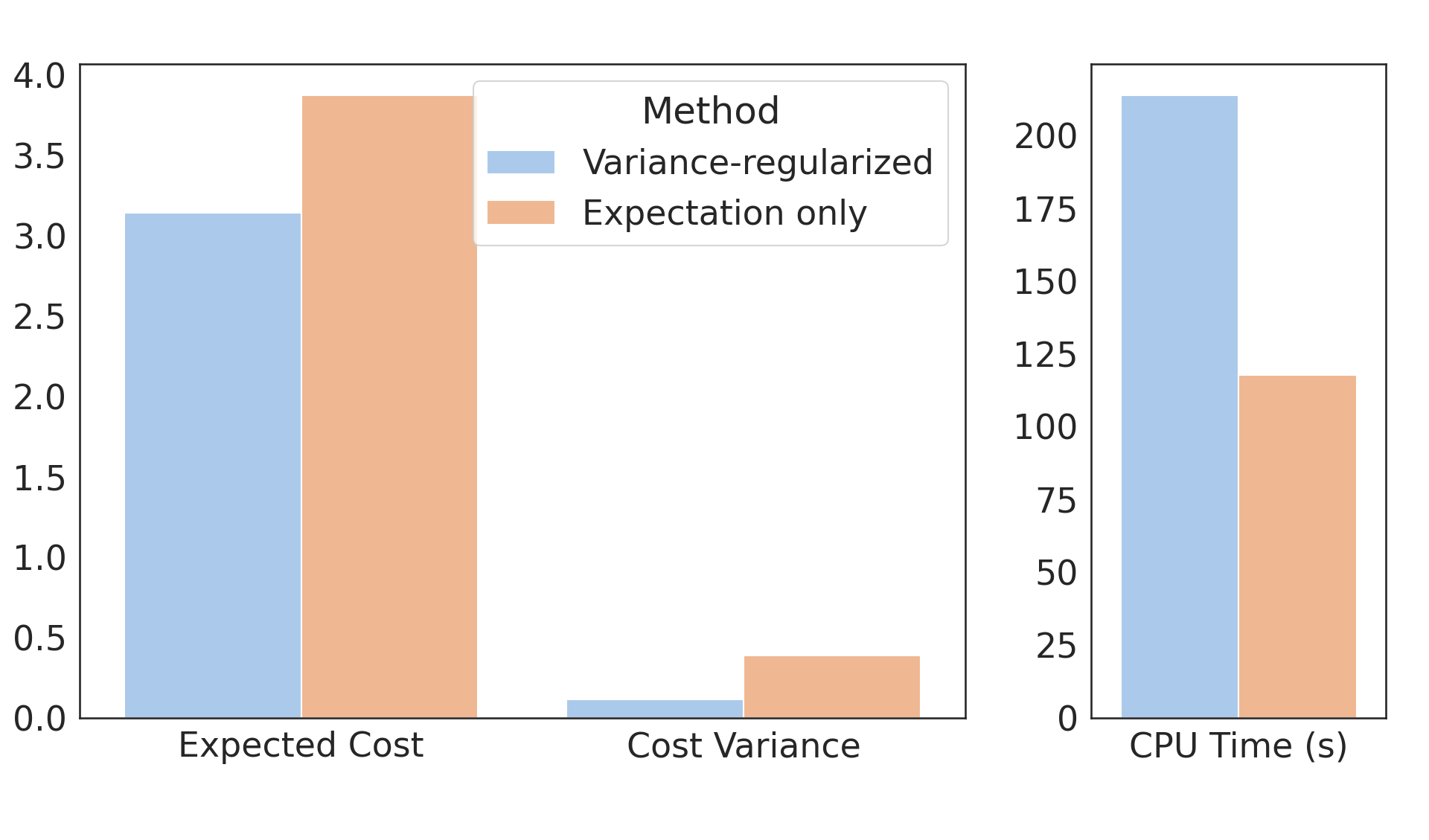}
        \caption{Effect of VR; sensor placement}
    \end{subfigure}%
    \begin{subfigure}[t]{0.25\linewidth}
        \centering
        \includegraphics[width=\linewidth]{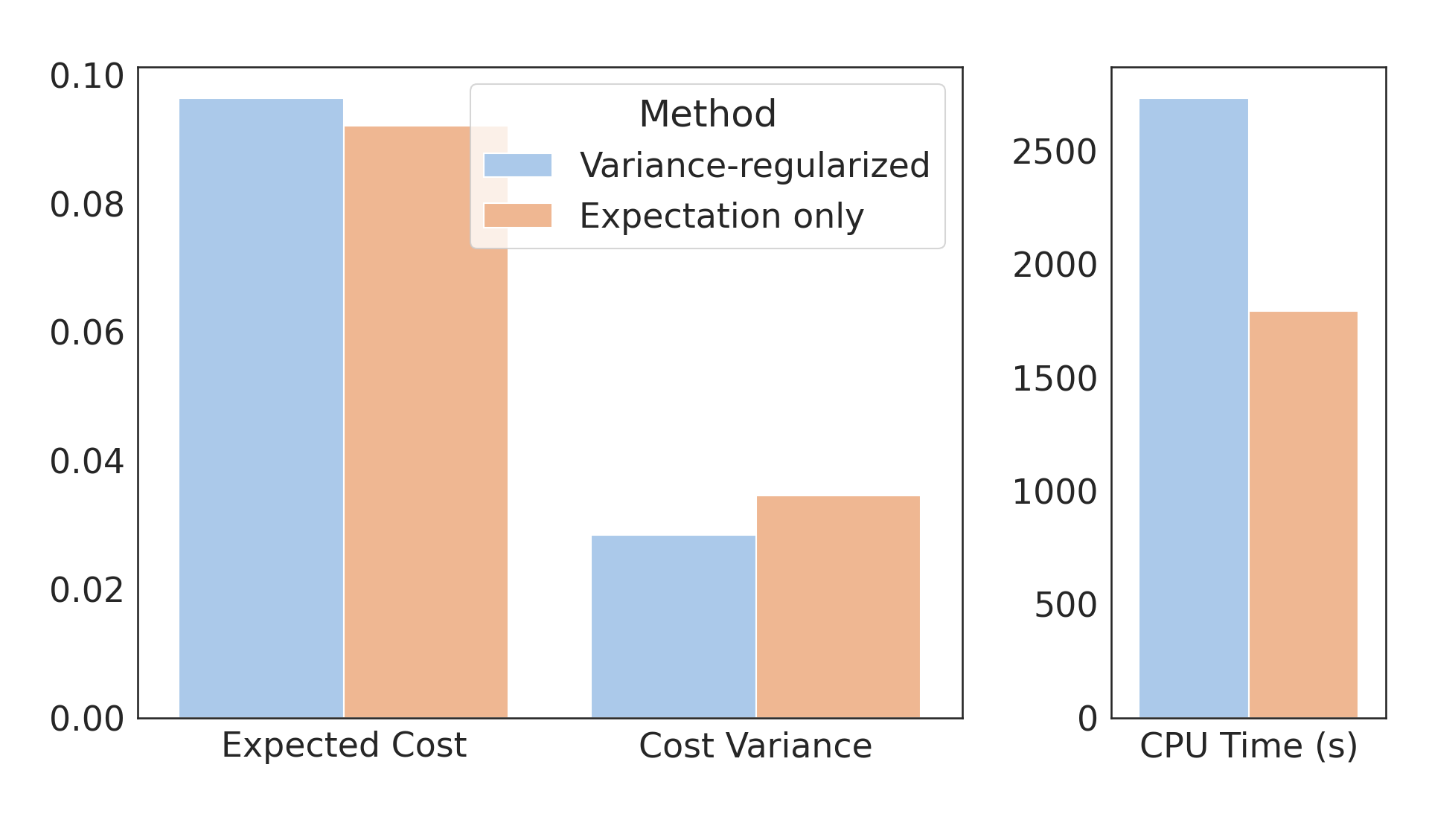}
        \caption{Effect of VR; manipulation}
    \end{subfigure}
    \caption{(a)-(b) Improvement of automatic differentiation (AD) over finite differences (FD) in both case studies. (c)-(d) Effect of variance regularization (VR) in both case studies.}
    \label{fig:ablation}
\end{figure*}

Our case studies in Sections~\ref{case1} and~\ref{case2} help demonstrate the utility of our framework for solving realistic robotics problems. However, it remains to justify the choices we made in designing this framework. For instance, how does automatic differentiation compare with other methods for estimating the gradient (e.g. finite differences)? What benefit does variance regularization in problem~\eqref{design_optimization_nlp} bring? We answer these questions here using an ablation study where we attempt to isolate the impact of each of these features.

First, why use automatic differentiation? On the one hand, AD allows us to estimate the gradient with only a single evaluation of the objective function, while other methods (such as finite differences, or FD) require multiple evaluations. On the other hand, AD necessarily incurs some overhead at runtime, making each AD function call more expensive than those used in an FD scheme. Additionally, some arguments~\cite{suh2021_bundled_gradients} suggest that exact gradients may be less useful than finite-difference or stochastic approximations when the objective is stiff or discontinuous. We compare AD with a 3-point finite-difference method by re-solving problem~\eqref{design_optimization_nlp} for both case studies, keeping all parameters constant ($N=512$, $\lambda=0.1$, same random seed) and substituting the gradients obtained using AD for those computed using finite differences. Fig.~\ref{fig:ablation} shows the results of this comparison. In the sensor placement example, AD achieves a lower expected cost and cost variance, and it runs in 32\% less time. In the collaborative manipulation example, both methods achieve similar expected cost and variance, but the AD version runs nearly 19x faster. These results lead us to conclude that AD enables more effective optimization than finite differences and is an appropriate choice for our framework. An exciting extension of our framework involves combining AD with stochastic population methods, but we leave this to future work.

The next question is whether variance regularization brings any benefit to the design optimization problem. To answer this question, we compare the results of re-solving both case studies with variance weight $\lambda = 0.1$ and $\lambda = 0$. These results are shown in Fig.~\ref{fig:ablation}; surprisingly, in the sensor placement example we see that the variance-regularized problem results in a lower expected cost, contrary to the intuition that regularization requires a trade off with increased expected cost. We expect that this lower expected cost may be a result of the regularization term smoothing the objective with respect to the exogenous parameters. However, these benefits are less pronounced than the benefits from automatic differentiation, and we do not see a distinct benefit in our second case study.


\subsection{Accuracy of robustness analysis}\label{soundness}

To verify the soundness of our statistical robustness analysis methods, we need to determine whether the fit GEVD is likely to either under- or overestimate the worst-case performance of a design. Put simply, is our approach falsely optimistic (underestimating the worst-case) or conservative (overestimating)?

To answer these questions, we compare the cumulative distribution function (CDF) of the fit GEVD with an empirical CDF observed from data. Algorithms~\ref{alg:worst_case_cost} and~\ref{alg:sensitivity} both estimate a posterior distribution for $\mu$, $\sigma$, and $\xi$, allowing us to construct an upper-bound and lower-bound GEVD using the 97\% and 3\% confidence level parameter estimates. Using these distributions, we can measure false optimism and conservatism using a one-sided Kolmogorov-Smirnov (KS) test~\cite{nist_ks}.

\begin{figure}[tb]
    \centering
    \includegraphics[width=\linewidth]{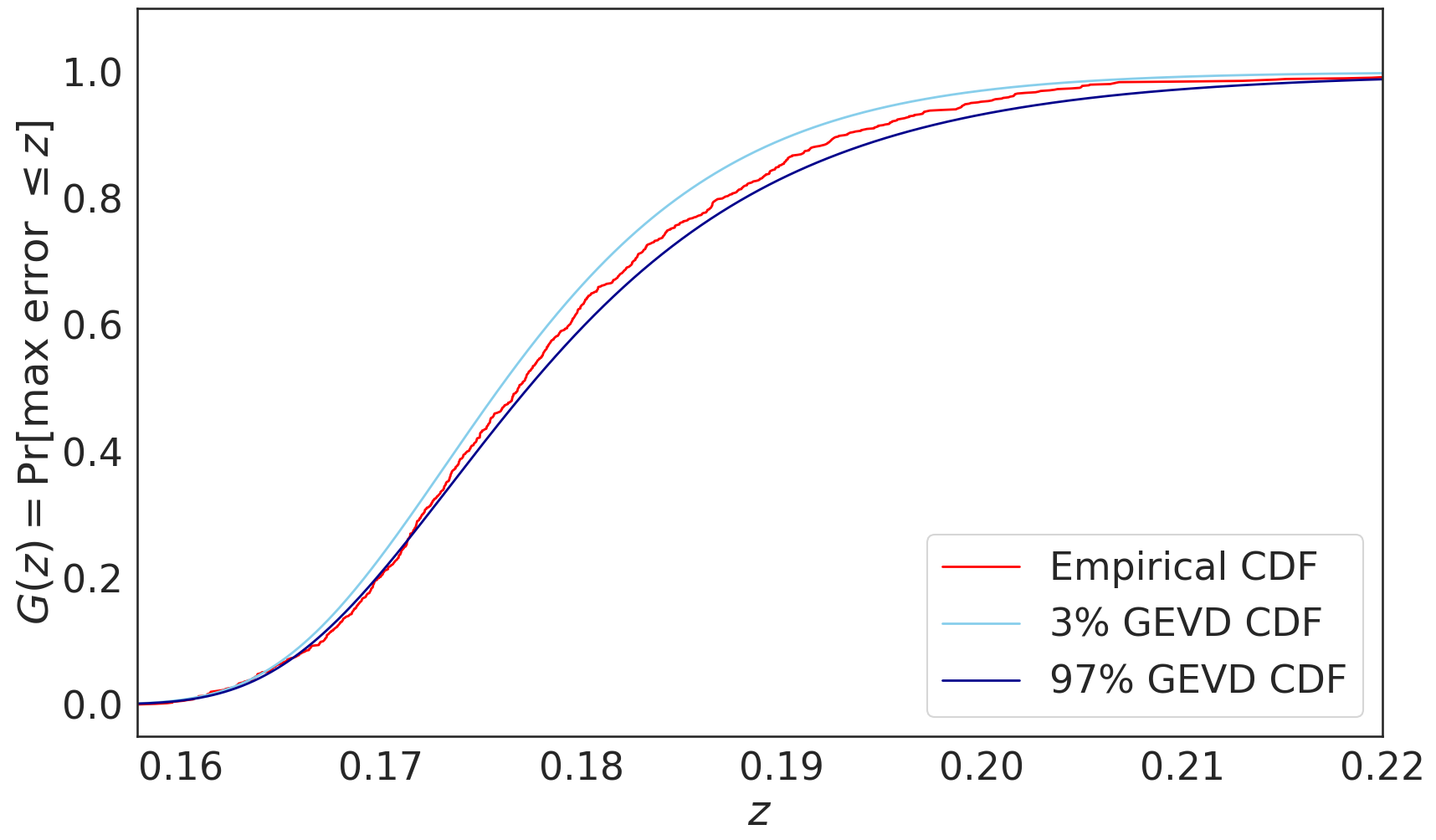}

    \includegraphics[width=0.99\linewidth]{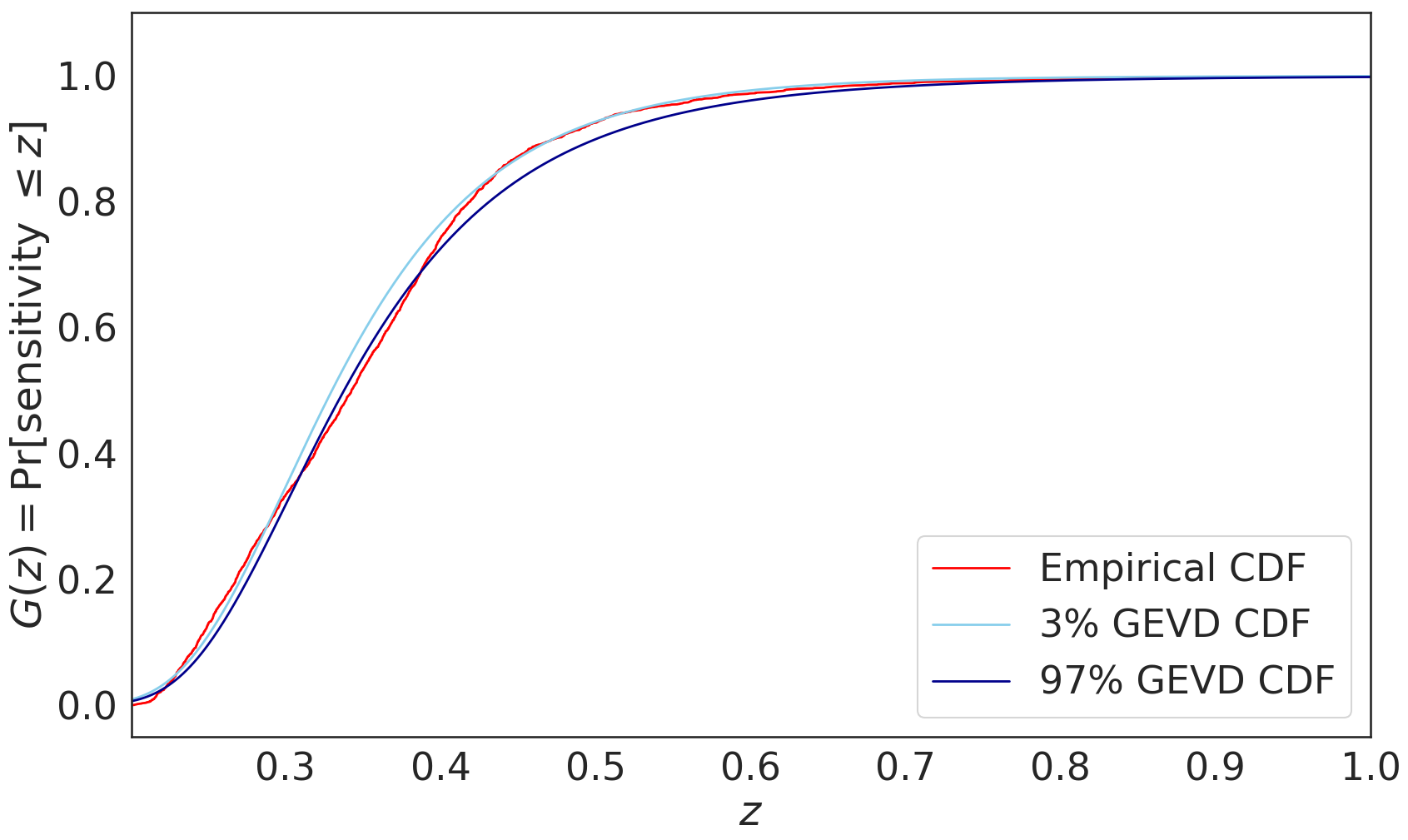}
    \caption{Comparison of fit GEVD CDFs and empirical CDF for worst-case estimation error in the sensor placement example (top) and sensitivity in the manipulation example (bottom).}
    \label{fig:ks_test}
\end{figure}

Fig.~\ref{fig:ks_test} compares the estimated GEVDs and empirical data for worst-case performance in the sensor placement example (fit using Algorithm~\ref{alg:worst_case_cost}) and sensitivity in the manipulation example (fit using Algorithm~\ref{alg:sensitivity}). In the former case, we see that the empirical CDF lies between the upper- and lower-confidence limits for the fit distribution, indicating that the fit is neither falsely optimistic at the 97\% level nor conservative at the 3\% level (these conclusions are confirmed by the KS statistics provided in Table~\ref{tab:ks_test_agv} in the appendix). In the latter case, even though the empirical CDF extends slightly beyond the estimated bounds in some regions, the statistical analysis in Table~\ref{tab:ks_test_mam} indicates that the estimated GEVD is neither falsely optimistic at the 97\% level nor conservative at the 3\% level. In addition, we see that the gap between the 3\% and 97\% distributions is relatively small in both examples in Fig.~\ref{fig:ks_test}.

\section{Discussion and Conclusion}

In this paper, we develop an automated design tool to improve the productivity of robot designers by a) enabling efficient optimization of robot designs and b) allowing users to certify the robustness of those designs. In developing this framework, we make two main algorithmic and theoretical contributions. First, we use differentiable programming for end-to-end optimization of robotic systems, creating a flexible software framework for design optimization. Second, we develop a novel statistical framework for certifying the worst-case performance and sensitivity of optimized designs.

To validate this framework and demonstrate the usefulness of our contributions, we present two case studies to highlight how our framework can be used for design optimization in practical robotics problems. Moreover, we show that our optimized designs are robust enough to deploy in hardware, and data from these hardware experiments validate our optimization approach. Finally, we provide an ablation study to justify the architecture of our optimization framework and a statistical analysis showing the soundness of our robustness analysis techniques. We hope that by combining flexible design optimization with robustness certification in our framework we can increase the productivity of robotics engineers, shorten the design cycle, and help bring more complex robotic systems to life.

There are a number of interesting directions for future work. First, since our approach relies on sampling from $\Phi$ without any further information, it will require a large number of samples to accurately capture rare events. We can close this gap when more information about $\Phi$ is available, perhaps using adversarial testing or importance sampling. Second, our framework is currently focused on tuning continuous parameters; we hope to incorporate stochastic search over discrete parameters in a future work. Finally, we hope to expand the software implementation of our framework to include a richer library of autonomy building blocks and demonstrate a wider range of applications in designing autonomous systems, including robotic arms, autonomous air and spacecraft, and networked autonomous systems. 

\bibliographystyle{IEEEtran}
\bibliography{IEEEabrv,main}

\appendix

\section*{Sensor Placement Design Problem Statement}

We model the robot with discrete-time Dubins dynamics with three state variables ($q = [x, y, \theta]$), two control inputs for linear and angular velocity ($u = [v, \omega]$), and noisy transition model
\begin{align*}
    \mat{x \\ y \\ \theta}_{t+1} = \mat{x \\ y \\ \theta}_{t} + \mat{\Delta t v \cos(\theta + \Delta t \omega / 2) \\ \Delta t v \sin(\theta + \Delta t \omega / 2) \\ \Delta t \omega} + w_t
\end{align*}
where $\Delta t = 0.5$ and $w_t \in \R^3$ is the actuation noise ($w_t \sim \mathcal{N}(0, Q)$ with covariance $Q \in \R^{3\times3}$). The measurement model is
\begin{align*}
    z_t = \mat{(x_t - x_{b1})^2 + (y_t - y_{b1})^2 \\ (x_t - x_{b2})^2 + (y_t - y_{b2})^2 \\ \theta} + v_t
\end{align*}
where $v_t$ is the measurement noise ($v_t \sim \mathcal(0, R)$ and covariance $R \in \R^{3\times3}$), modeling range measurements from radio or acoustic beacons $b_1$ and $b_2$ and inertial or magnetic measurements of $\theta$. The initial state of the robot is normally distributed $q_0 \sim \mathcal{N}(\bar{q}_0, P_0)$ for mean initial state $\bar{q}_0 \in \R^3$ and initial covariance $P_0 \in \R^{3\times3}$. The navigation function (shown in Fig.~\ref{fig:agv_representative_trajectories}) is $V_t(x_t, y_t) = 2 (x_t^2 + y_t^2) + 0.05 / d_t$ ($d_t$ is the distance from the robot to the nearest obstacle at step $t$). Formally, we define this problem in the language of our framework in Table~\ref{tab:agv_design_problem}.

\begin{table}[]
\renewcommand{\arraystretch}{1.5}
\caption{Formal statement of the sensor placement design problem with $T$ discrete timesteps.}
\label{tab:agv_design_problem}
\begin{tabular}{|p{1.25cm}|p{7cm}|}
\hline
Design\ \ parameters & \begin{tabular}[c]{@{}l@{}}$\theta = [b_1, b_2, k] \in \R^6$\\ Beacon locations: $b_i = (x_{bi}, y_{bi}) \in \R^2$ for $i=1,2$\\ Feedback gains: $k \in \R^2$\end{tabular} \\ \hline
Exogenous parameters & \begin{tabular}[c]{@{}l@{}}$\phi = [q_0, w_0, \ldots, w_{T-1}, v_0, \ldots, v_{T-1}] \in \R^{3 + 6T}$\\ Initial state: $q_0 \in \R^3,\ q_0 \sim \mathcal{N}(\bar{q}_0, P_0)$;\\\phantom{Initial State: }$P_0 = 0.001 I_{3\times3}$\\ Actuation noise: $w_t \in \R^3,\ w_t\sim \mathcal{N}(0, Q)$;\\\phantom{Actuation noise: }$Q = (\Delta t)^2 \text{diag}\pn{[0.001, 0.001, 0.01]}$\\ Measurement noise: $v_t \in \R^3,\ v_t\sim \mathcal{N}(0, R)$\\\phantom{measurement noise: }$R = \text{diag}\pn{[0.1, 0.01, 0.01]}$\end{tabular} \\ \hline
Simulator & $S$ initializes the robot with state $q_0$ and EKF state estimate $\bar{q}_0$ and error covariance $P_0$, then steps forward with interval $\Delta t = 0.5$ for $T = 60$ total steps. At each step, the simulator
\begin{enumerate}
    \item Evaluates the navigation function to find a collision-free path to the goal,
    \item Uses a feedback controller to track that path,
    \item Updates the state using forward Euler integration,
    \item Performs an EKF prediction, obtains a measurement $z_t$, and performs an EKF update.
\end{enumerate}

$S$ returns a trace $s_t = [q, \hat{q}, P_{t|t}, V_t]$ containing true states, estimated states, estimated posterior error covariance, and the value of the navigation function at each time step. \\ \hline
Cost & $J$ has three components. The first ($\norm{q_t - \hat{q}_t}^2$) minimizes the estimation error of the EKF, the second ($\norm{q_t}$) guides the robot towards the goal, and the third (both $V_t$ terms) avoids collision with the environment:
$J = \frac{1}{T} \sum_{t=1}^T \pn{100 \norm{q_t - \hat{q}_t}^2 + \norm{q_t}^2 + 0.1 V_t}$ $+ 0.1 \max_t V_t$ \\ \hline
Constraints & $(x_{bi}, y_{bi}) \in [-3, 0] \times [-1, 1]$ for $i=1, 2$ \\ \hline
\end{tabular}
\end{table}

\section*{Multi-agent Manipulation Design Problem Statement}

We model each ground robot as a double integrator with states $[p_x, p_y, \theta, v_x, v_y, \omega]$. Given control inputs representing desired linear velocity $v_d$ in the $[\cos\theta, \sin\theta]$ direction and desired angular velocity $\omega_d$, the robot tracks those desired velocities by applying forces and torques subject to a friction cone constraint. The box is modeled as a rigid body with friction against the ground. Contact forces between the box and each robot are modeled using a penalty method described in~\cite{suh2021_bundled_gradients}, where the normal force is given by $f_n = k_c \min(\phi, 0) - k_d \dot{\phi} \mathbbm{1}_{\phi < 0}$ ($\phi$ is the signed distance between the robot and the box, $k_c = \SI{300}{N/m}$ is the contact stiffness, $k_d$ is a damping coefficient chosen to ensure critical damping, and $\mathbbm{1}_{\phi < 0}$ is the indicator function equal to 1 when the box and robot are in contact and 0 otherwise). Friction in the box/ground and box/robot contacts was modeled as Coulomb friction, resulting in a tangential force $f_t = \mu f_n$ with $\mu = c\psi$ if $\psi < \psi_s$ and $\mu = \mu_d$ otherwise, where $mu_d$ is the coefficient of dynamic friction ($\mu_d$ varies for each contact pair), $\psi$ is the tangential velocity at the point of contact, $\psi_s = \SI{0.3}{m/s}$ is the tangential velocity where slipping begins, and $c = \mu_d / \psi_s$ was chosen to ensure a continuous friction model.

Each ground robot uses a proportional controller (with tunable gains) to find $v_d$ and $\omega_d$ to track a cubic spline reference trajectory. The start point of each spline is set to match the robot's current position, the end point is set based a known offset from the desired box location, and the central control point of the spline is set using a neural network (with tunable parameters). The neural network is given inputs including the current position of each robot and the desired box pose, all referenced against the current box pose, and it predicts $(x, y)$ locations for the control point for each robot. The network uses $\tanh$ activations on each hidden layer.

Formally, we define this problem in the language of our framework (design parameters, exogenous parameters, etc.) in Table~\ref{tab:mam_design_problem}. The design parameters include the trajectory tracking control gains and network parameters, while the exogenous parameters include the desired box pose, coefficients of friction, box mass, and initial robot poses.

\begin{table}[]
\renewcommand{\arraystretch}{1.5}
\caption{Formal statement of the collaborative manipulation design problem using a planning network with $n_p$ total parameters (weights and biases).}
\label{tab:mam_design_problem}
\begin{tabular}{|p{1.25cm}|p{7cm}|}
\hline
Design\ \ parameters & \begin{tabular}[c]{@{}l@{}}$\theta = [k_v, k_\omega, w_i, b_i] \in \R^{2 + n_p}$\\ Trajectory tracking gains: $[k_v, k_w] \in \R^2$\\ Network weights and biases: $(w_i, b_i)$ for $i=1,\ldots,n_p$\end{tabular} \\ \hline
Exogenous parameters & \begin{tabular}[c]{@{}l@{}}$\phi = [\mu_{rg}, \mu_{bg}, \mu_{br}, m_b, p_{bd}, p_{r1}, p_{r2}] \in \R^{13}$\\ Robot/ground, box/ground, box/robot coefficients of friction:\\\quad$[\mu_{rg}, \mu_{bg}, \mu_{br}] \in [0.6, 0.8] \times [0.4, 0.6] \times [0.1, 0.3]$ \\
Box mass: $m_b \in [0.9, 1.1]$ \\
Desired box pose: \\\quad $p_{bd} = [x_d, y_d, \theta_d] \in [0, 0.5]^2 \times [-\pi/4, \pi/4]$ \\ (Above parameters are uniformly distributed) \\
Initial robot pose: $p_{ri} = [x_{0}, y_0, \theta_0] \sim \mathcal{N}(\bar{p}_{ri}, \Sigma)$;\\\quad $\Sigma = 0.01 I_{3\times3}$, $i=1,2$. \end{tabular} \\ \hline
Simulator & $S$ initializes the robots at the initial states in $\phi$ relative to the box. Since these initial states may be in contact, we simulate \SI{0.5}{s} of settling time at a \SI{0.01}{s} timestep, then re-index the robot positions and desired box pose relative to the settled box pose. We then evaluate the planning network and track the planned path for \SI{4}{s} at a \SI{0.01}{s} timestep. At each timestep, 1) evaluate the spline tracking controller, 2) evaluate contact dynamics between the box, robots, and ground, and 3) integrate forces and torques to obtain box and robot states at the next timestep. $S$ returns a trace $s_t = [q_{r1}, q_{r2}, q_{b}]$ containing the states of each robot and the box over time (relative to the initial pose of the box after the settling period). \\ \hline
Cost & $J$ is simply the squared distance between the final box pose and the desired box position $(x - x_d)^2 + (y - y_d)^2 + (\theta - \theta_d)^2$ \\ \hline
Constraints & Network parameters were not constrained. $k_v$ and $k_w$ were constrained to be less than 10. \\ \hline
\end{tabular}
\end{table}

\begin{figure*}[t]
    \centering
    \includegraphics[width=\linewidth]{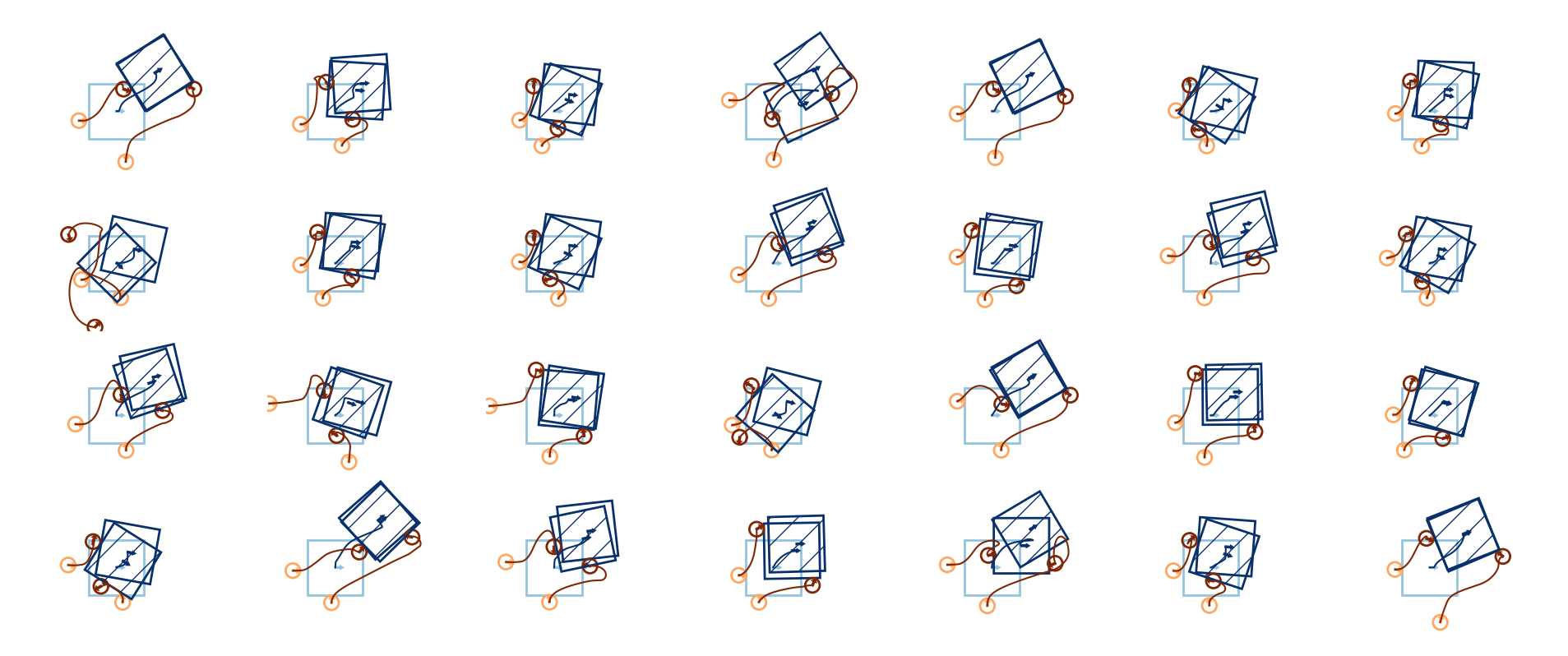}
    \caption{Additional examples of optimized multi-agent manipulation behavior in simulation, showing that the optimized strategy reaches the goal in most cases. Each example shows the results of executing the optimized pushing strategy for \SI{4}{s} with a randomly selected set of friction coefficients, random target pose, and random initial robot poses. Light/dark colors indicate initial/final positions, respectively, and the striped box indicates the target pose.}
    \label{fig:mam_more}
\end{figure*}

\section*{Kolmogorov-Smirnov Test Results}

Table~\ref{tab:ks_test_agv} provides results from one-sided KS tests for the GEVD estimated using Algorithm~\ref{alg:worst_case_cost} in the sensor placement case study, while Table~\ref{tab:ks_test_mam} provides similar results for Algorithm~\ref{alg:sensitivity} in the collaborative manipulation case study.

\begin{table*}[thb]
\renewcommand{\arraystretch}{1.5}
\centering
\begin{tabular}{p{1.5cm}||p{6cm}|c|c|p{6cm}}
 & Null Hypothesis & KS Statistic & p-value & Conclusion \newline ($p < 0.05$) \\ \hline\hline
False Optimism & 97\% GEVD under-estimates worst-case performance & 0.0410 & 0.0337 & Reject; 97\% GEVD \textit{does not} under-estimate worst-case performance \\ \hline
Conservatism & 3\% GEVD over-estimates worst-case performance & 0.0529 & 0.00354 & Reject; 3\% GEVD \textit{does not} over-estimate worst-case performance
\end{tabular}
\caption{Results of one-sided KS tests for the sensor placement case study. These results indicate that Algorithm~\ref{alg:worst_case_cost} is sound in this case.}\label{tab:ks_test_agv}
\end{table*}

\begin{table*}[thb]
\renewcommand{\arraystretch}{1.5}
\centering
\begin{tabular}{p{1.5cm}||p{6cm}|c|c|p{6cm}}
 & Null Hypothesis & KS Statistic & p-value & Conclusion \newline ($p < 0.05$) \\ \hline\hline
False Optimism & 97\% GEVD under-estimates sensitivity & 0.0399 & $6.75\times10^{-5}$ & Reject; 97\% GEVD \textit{does not} under-estimate sensitivity \\ \hline
Conservatism & 3\% GEVD over-estimates sensitivity & 0.0618 & $1.03\times10^{-10}$ & Reject; 3\% GEVD \textit{does not} over-estimate sensitivity
\end{tabular}
\caption{Results of one-sided KS tests for the collaborative manipulation case study. These results indicate that Algorithm~\ref{alg:sensitivity} is sound in this case.}\label{tab:ks_test_mam}
\end{table*}

\end{document}